\documentclass{article} 
\usepackage{iclr2025_conference,times}

\usepackage{amsmath,amsfonts,bm}

\def\eqref#1{equation~\ref{#1}}

\def\1{\bm{1}}

\def\rvf{{\mathbf{f}}}

\def\rvp{{\mathbf{p}}}

\def\rvx{{\mathbf{x}}}

\def\rvz{{\mathbf{z}}}

\def\rmR{{\mathbf{R}}}

\def\rmX{{\mathbf{X}}}

\DeclareMathAlphabet{\mathsfit}{\encodingdefault}{\sfdefault}{m}{sl}
\SetMathAlphabet{\mathsfit}{bold}{\encodingdefault}{\sfdefault}{bx}{n}

\def\gC{{\mathcal{C}}}

\def\gF{{\mathcal{F}}}

\def\gL{{\mathcal{L}}}

\def\gO{{\mathcal{O}}}
\def\gP{{\mathcal{P}}}
\def\gQ{{\mathcal{Q}}}

\def\gX{{\mathcal{X}}}

\def\gZ{{\mathcal{Z}}}

\def\sR{{\mathbb{R}}}

\definecolor{cvprblue}{rgb}{0.21,0.49,0.74}
\usepackage[breaklinks,colorlinks,citecolor=cvprblue]{hyperref}
\usepackage{url}

\usepackage{comment}
\usepackage[capitalize]{cleveref}
\usepackage{svg}
\usepackage{overpic}
\usepackage{tikz}
\usepackage{pgfplots}
\usetikzlibrary{positioning,calc,fadings,backgrounds,fit,3d,shapes.misc,shapes.geometric,matrix,hobby}
\pgfplotsset{compat=1.14}
\usepackage{wrapfig}
\usepackage[fixed]{fontawesome5}
\usepackage{colortbl}
\usepackage{booktabs}
\usepackage{multirow}
\definecolor{mytbcol}{RGB}{175,227,246}
\definecolor{mytitlecol}{RGB}{102,102,255}

\usepackage{listings}

\usepackage{array}
\newcolumntype{H}{>{\setbox0=\hbox\bgroup}c<{\egroup}@{}}

\definecolor{codegreen}{rgb}{0,0.6,0}
\definecolor{codegray}{rgb}{0.5,0.5,0.5}
\definecolor{codepurple}{rgb}{0.58,0,0.82}
\definecolor{backcolour}{rgb}{0.95,0.95,0.92}

\lstdefinestyle{mystyle}{
    backgroundcolor=\color{backcolour},   
    commentstyle=\color{codegreen},
    keywordstyle=\color{magenta},
    numberstyle=\tiny\color{codegray},
    stringstyle=\color{codepurple},
    basicstyle=\ttfamily\footnotesize,
    breakatwhitespace=false,         
    breaklines=true,                 
    captionpos=b,                    
    keepspaces=true,                 
    numbers=left,                    
    numbersep=5pt,                  
    showspaces=false,                
    showstringspaces=false,
    showtabs=false,                  
    tabsize=2
}
\iclrfinalcopy
\title{\textcolor{mytitlecol}{LaGeM\faGem[regular]}: A \textcolor{mytitlecol}{La}rge \textcolor{mytitlecol}{Ge}ometry \textcolor{mytitlecol}{M}odel for 3D Representation Learning and Diffusion}

\author{Biao Zhang, Peter Wonka\\
KAUST, Saudi Arabia \\
\texttt{\{biao.zhang, peter.wonka\}@kaust.edu.sa}
}

\begin{document}

\maketitle

\begin{abstract}
This paper introduces a novel hierarchical autoencoder that maps 3D models into a highly compressed latent space. The hierarchical autoencoder is specifically designed to tackle the challenges arising from large-scale datasets and generative modeling using diffusion. Different from previous approaches that only work on a regular image or volume grid, our hierarchical autoencoder operates on unordered sets of vectors. Each level of the autoencoder controls different geometric levels of detail. We show that the model can be used to represent a wide range of 3D models while faithfully representing high-resolution geometry details. The training of the new architecture takes 0.70x time and 0.58x memory compared to the baseline.
We also explore how the new representation can be used for generative modeling. Specifically, we propose a cascaded diffusion framework where each stage is conditioned on the previous stage. Our design extends existing cascaded designs for image and volume grids to vector sets.
\end{abstract}

\section{Introduction}

Diffusion models are currently the best-performing models for image, video, and 3D object generation.
For 3D object generation, there are two main branches of research. The first branch, pioneered by Dreamfusion~\citep{poole2022dreamfusion}, aims to lift 2D diffusion models to 3D model generation. The advantage of this method is that it can benefit from the large-scale 2D datasets used for training 2D diffusion models and it sparked a lot of follow-up work~\citep{poole2022dreamfusion, wang2023score, lin2023magic3d, chen2023fantasia3d, wang2024prolificdreamer, qian2023magic123, tang2023dreamgaussian, yi2023gaussiandreamer, wang2023imagedream, liu2024one, long2024wonder3d, zheng2024mvd, li2023instant3d, ho2022cascaded, xu2023dmv3d}.
The second branch tackles the training on 3D datasets directly. The advantage of this method is that it is more direct and leads to faster inference times~\citep{mittal2022autosdf, yan2022shapeformer, zhang20223dilg, zeng2022lion, zheng2023locally, hui2022neural, vecset, siddiqui2024meshgpt, chen2024meshxl, chen2024meshanything}. Our work sets out to contribute to this second branch of methods.

Among these 3D native generation methods, 3DShape2VecSet~\citep{vecset} (or VecSet for short) has been proven to be an effective method to encode 3D geometry. It proposed an autoencoder to find an efficient representation for 3D models as a set of vectors. Because of the high reconstruction quality and compactness of the latent space, the method alleviates the difficulty of training 3D generative models. Some other works~\citep{zhao2024michelangelo, cao2024motion2vecsets, dong2024gpld3d, petrov2024gem3d, zhang2024clay, zhang2024functional} follow the VecSet representation. 
We noticed that VecSet's expressiveness is limited by the number of latent vectors. It is overfitting on smaller datasets like ShapeNet and is unable to scale to larger datasets. To improve the expressiveness, we need to scale up the latent size and the training dataset. The straightforward way is to employ hundreds of GPUs for training which is expensive~\citep{zhang2024clay}. Thus, our goal is to reduce the training cost in terms of time and memory consumption while achieving similar or even better autoencoding quality.

\begin{figure}[htb]
    \centering
    \input{tex/ae}
    \vspace{-20pt}
    \caption{\textbf{Autoencoders.} We show different autoencoder architectures here, including AE (AutoEncoder), U-Net, VAE~\citep{kingma2013auto}, NVAE~\citep{vahdat2020nvae}, VecSet~\citep{vecset} and the proposed LaGeM. VAE and NVAE are for image data, while VecSet and LaGeM are for geometry (distance function) data. In the top row, VAE and VecSet are using a single scale latent to represent the data. Both NVAE and LaGeM use multi-scale latents to represent data. All the previous works VAE, NVAE, and VecSet apply KL divergence in the bottleneck to regularize the latent space, while in this work, we apply standardization in the bottleneck.}
    \label{fig:ae}
\end{figure}

In the image domain, NVAE~\citep{vahdat2020nvae} extended the design of the variational autoencoder (VAE)~\citep{kingma2013auto} to a hierarchical VAE based on the design of the U-Net. The latent space of the NVAE is a multi-scale latent grid and the reconstruction quality of the images from the NVAE improves a lot over the VAE. An illustration of the architectures can be found in~\cref{fig:ae}.
We draw inspiration from the design of the NVAE and design a multi-scale latent VecSet representation, called \emph{LaGeM}. We train our architecture on a large-scale geometry dataset Objaverse~\citep{deitke2023objaverse} and improve training time by 0.7 and memory consumption by 0.58 compared to VecSet. 

\begin{wraptable}{r}{3.5cm}
\vspace{-25pt}
\begin{tabular}{cc}\\\toprule  
Latents & Controlling \\ \midrule
Level 3 & Main Structure \\
Level 2 & Major Details \\
Level 1 & Minor Details \\
\bottomrule
\end{tabular}
\end{wraptable}
Additionally, we also propose a cascaded generative model for the hierarchical latent space. We generate the latent VecSet from the lower resolution level to the highest resolution level stage-by-stage. In each stage, we use the previously generated latents as conditioning information. As a result, this enables control over the level of detail of the generated geometry. 

We summarize our contributions as follows:
\begin{itemize}\setlength\itemsep{-0.1em}
    \item We propose a hierarchical autoencoder architecture with faster training time and low memory consumption. The latent space is composed of several levels.
    \item The model is capable of training on large-scale datasets like objaverse. 
    \item We propose a cascaded diffusion model to generate 3D geometry in the hierarchical latent space. This enables control of the level of detail of the generated model.
\end{itemize}
\begin{table}[]
    \centering
    \caption{\textbf{Geometric Latent Representation and Generation.}}
\def\arraystretch{1.15}\tabcolsep=0.32em
\begin{tabular}{cHccc@{\hspace*{-\tabcolsep}}}
\toprule
Method            & Venue              & Learning Method           &    Latent Rep            &  Hierarchies      \\ \midrule
\rowcolor{mytbcol!30}ShapeFormer~\citep{yan2022shapeformer}       & CVPR 2022          & AutoEncoder & Sparse Volume   & Single \\
3DILG~\citep{zhang20223dilg}             & NeurIPS 2022       & AutoEncoder & Irregular Grid & Single \\
\rowcolor{mytbcol!30}LION~\citep{zeng2022lion} & NeurIPS 2022 & AutoEncoder & Latent Points & Multi \\
TriplaneDiffusion~\citep{shue20233d} & CVPR 2023          & AutoDeocder & Planes       & Single \\
\rowcolor{mytbcol!30}SDFusion~\citep{cheng2023sdfusion}          & CVPR 2023          & AutoEncoder & Volume          & Single \\
3DShape2VecSet~\citep{vecset}    & SG 2023      & AutoEncoder & VecSet         & Single \\
\rowcolor{mytbcol!30}HyperDiffusion~\citep{erkocc2023hyperdiffusion}    & ICCV 2023          & Per-Object Optimization  & Network Weight         & Single \\
XCube~\citep{ren2024xcube}             & CVPR 2024          & AutoEncoder & Sparse Volume   & Multi  \\
\rowcolor{mytbcol!30}Mosaic-SDF~\citep{yariv2024mosaic}       & CVPR 2024          & Per-Object Optimization  & Irregular Grid & Single \\
3DTopia-XL\citep{chen20243dtopia} & & Per-Object Optimization & Irregular Grid & Single\\
\rowcolor{mytbcol!30}OctFusion~\citep{xiong2024octfusion}         & arxiv 2024         & AutoEncoder & Sparse Volume   & Multi  \\ 
\midrule
\textbf{LaGeM}\faGem[regular] (Ours)             &                    & AutoEncoder & VecSet         & Multi  \\ \bottomrule
\end{tabular}
    \label{tab:geo_gen}
\end{table}
\section{Related works}
We show an overview of latent 3D generative models in~\cref{tab:geo_gen}, particularly focusing on the type of latent space used.
\subsection{Learning Methods}
Usually, a learning method is required to convert 3D geometry to latent space. 1) One way to do this is to convert 3d geometry to latent space with a per-object optimization method, e.g.~\citep{erkocc2023hyperdiffusion, yariv2024mosaic}. For larger datasets, this approach is very time-consuming. 2) Alternatively, auto-decoder, e.g., DeepSDF~\citep{park2019deepsdf}, jointly optimize the latent space for all objects in the dataset. However, as there is no encoder, new objects cannot be mapped to latent space easily. 3) Therefore, a commonly used framework is the auto-encoder. The optimization is efficient because it is performed jointly for all objects in the dataset, and new objects not in the training set can be quickly encoded using the encoder. Thus, we also build on this approach. 

\subsection{Latent Representations}
Early methods used regular grids~\citep{yan2022shapeformer, cheng2023sdfusion} as the latent representation because of their simple structure. We can easily use convolutional layers to process volume data. To represent high-quality geometric details, we need large-resolution volumes. This makes the training even more difficult because of the $O(n^3)$ complexity. A way to solve this problem is to introduce sparsity~\citep{ren2024xcube} to the representation like octrees~\citep{xiong2024octfusion} or sparse irregular grids~\citep{zhang20223dilg, yariv2024mosaic}. Both structures have the potential to represent high-quality 3D models, but generating irregular structures explicitly is difficult for diffusion models. Different from the above mentioned approaches, 3DShape2VecSet~\citep{vecset} is proposed to solve the reconstruction problem without using any sparse structures. The representation is easy to use. In this paper, we investigate how to improve the VecSet representation. Compared to \citet{vecset}, our goal is to obtain an even higher-quality latent space by introducing Level of Latents (LoL).

\subsection{Cascaded Generation}
In the field of image generation, there are multiple cascaded diffusion models,e.g.,~\citep{ho2022cascaded, saharia2022photorealistic}. In the 3D domain, some works~\citep{zeng2022lion, ren2024xcube} also modeled geometries with hierarchical latents and proposed 3D generative models using cascaded diffusion models. Our work encodes 3D geometry into hierarchical VecSets. Thus, it is straightforward to consider cascaded latent diffusion to train generative models in our latent space.



\begin{figure}
    \centering
    \input{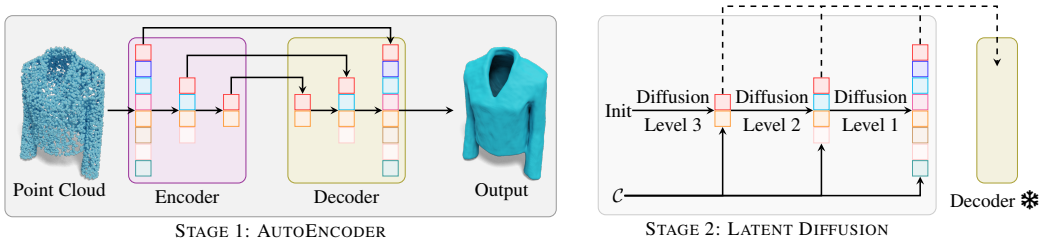}
    \vspace{-20pt}
    \caption{\textbf{Pipeline.} We proposed a U-Net-style transformer for the autoencoding. In this way, we obtain a hierarchical latent space, which contains several levels of latents. To train the generative diffusion models in the latent space, we propose the cascaded latent diffusion models.}
    \label{fig:pipeline}
\end{figure}

\section{Methodology}
\subsection{Background of VecSet Representations}
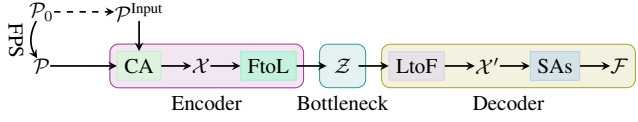
\begin{figure}
    \centering
\scalebox{0.8}{
\begin{tikzpicture}[]



    \tikzstyle{encoder} = [rectangle, rounded corners, draw=violet!80, fill=violet!10]
    \tikzstyle{decoder} = [rectangle, rounded corners, draw=olive!80, fill=olive!10]
    \tikzstyle{bottleneck} = [rectangle, rounded corners, draw=teal!80, fill=teal!10]

    \pgfdeclarelayer{background}
    \pgfdeclarelayer{box}
    \pgfsetlayers{box, background,main}
    
    \definecolor{mycol1}{RGB}{187,230,191}
    \definecolor{mycol2}{RGB}{170,197,209}
    \definecolor{mycol3}{RGB}{137,230,191}
    \definecolor{mycol4}{RGB}{210,197,209}
    
    \tikzstyle{arrow} = [thick,->,>=stealth]

    \tikzstyle{cross} = [rectangle, draw=mycol1!80, fill=mycol1!50]
    \tikzstyle{self} = [rectangle, draw=mycol2!80, fill=mycol2!50]
    \tikzstyle{ftl} = [rectangle, draw=mycol3!80, fill=mycol3!50]
    \tikzstyle{ltf} = [rectangle, draw=mycol4!80, fill=mycol4!50]    

    \node[inner sep=0pt,] (P0) {$\gP_0$};
    \node[right=1cm of P0, inner sep=0pt,] (P01) {$\gP^{\text{Input}}$};
    
    \node[cross, below=0.5cm of P01] (CA1) {CA};
    \node[inner sep=0pt,] at (CA1 -| P0) (P1) {$\gP$};

    \node[right=0.5cm of CA1, inner sep=0pt,] (X1) {$\gX$};
    
    \node[ftl, right=0.5cm of X1]  (FTL1) {FtoL};

    \node[right=0.5cm of FTL1] (Z1) {$\gZ$};

    \node[ltf, right=0.5cm of Z1] (LTF1) {LtoF};

    \node[right=0.5cm of LTF1, inner sep=0pt,] (X11) {$\gX'$};

    \node[self, right=0.5cm of X11,] (SA1) {SAs};

    \node[right=0.5cm of SA1, inner sep=0pt,] (F1) {$\gF$};

    \draw [arrow] (P01) -- (CA1);
    
    \draw [arrow] (P1) -- (CA1);
    \draw [arrow] (CA1) -- (X1);
    
    \draw [arrow] (X1) -- (FTL1);
    \draw [arrow] (FTL1) -- (Z1);
    \draw [arrow] (Z1) -- (LTF1);
    \draw [arrow] (LTF1) -- (X11);

    \draw [arrow] (X11) -- (SA1);
    \draw [arrow] (SA1) -- (F1);

    \draw [arrow, dashed] (P0) -- (P01);

    \draw [arrow] (P0) to [bend right=30]  node [below, sloped]  () {FPS} (P1);

    \coordinate[] (encode_lower_coord) at (Z1.south -| X1);
    
    \begin{pgfonlayer}{background}
        \node [draw, encoder, fit=(CA1)(FTL1)(encode_lower_coord), label={below:Encoder}] {}; 
    \end{pgfonlayer}

    \coordinate[] (decode_lower_coord) at (Z1.south -| F1);
    
   \begin{pgfonlayer}{background}
        \node [draw, decoder, fit=(LTF1)(F1)(decode_lower_coord), label={below:Decoder}] {}; 
    \end{pgfonlayer}

   \begin{pgfonlayer}{background}
        \node [draw, bottleneck, fit=(Z1), label={below:Bottleneck}] {}; 
    \end{pgfonlayer}
    
\end{tikzpicture}
}
    \vspace{-10pt}
    \caption{\textbf{Geometry Autoencoder.} The design from VecSet~\citep{vecset} can be seen as a special case of the proposed LaGeM network with only one level.}
    \label{fig:lgm1}
\end{figure}

The VecSet~\citep{vecset} representation converts a dense point cloud to a latent vector set $\gZ=\{\rvz_1, \rvz_2, \dots, \rvz_M\}$ with $\rvz\in\sR^D$ so that an occupancy/distance function $\gO(\rvp)$ can be recovered from the vector set. The simplified network is illustrated in~\cref{fig:lgm1}.
\paragraph{Encoding.} The process first downsamples the 3D input point cloud $\gP^{\text{Input}}=\{\rvp_i\}_{i=1,\dots, N}$ with furthest point sampling (FPS), $\gP=\mathrm{FPS}(\gP^{\text{Input}}, r)$,
where $r$ is the down-sampling ratio, and $\gP$ is a low-resolution version of $\gP^{\text{Input}}$. Then $\gP^{\text{Input}}$ is converted to an unordered set with cross-attention
\begin{equation}
    \mathrm{CA}(Q=\mathrm{PE}(\gP), K=\mathrm{PE}(\gP^{\text{Input}}), V=\mathrm{PE}(\gP^{\text{Input}})) = \gX = \{\rvx\in\sR^C\}_{i=1,2,\dots, M},
\end{equation}
where PE is a positional embedding function~\citep{vecset} and $\mathrm{CA}(\cdot, \cdot, \cdot)$ is a cross-attention module. We also write $\mathrm{CA}(\gP, \gP^{\text{Input}})$ for short.
Here, the positional embedding used to project a 3D coordinate $\rvp\in\sR^3$ to the high dimensional space $\sR^C$ is omitted for simplicity.
To obtain a highly compressed latent space, the vectors in $\gX$ are further compressed to a lower-dimensional space $\sR^D$ where $D\leq C$ (Feature to Latent, or FtoL in short),
\begin{equation}
    \mathrm{FtoL}(\gX) = \gZ=\{\rvz\in\sR^D\}_{i=1,2,\dots, M}.
\end{equation}
This compression step is also regularized by KL divergence.
\paragraph{Decoding.} Each latent vector in $\gZ$ is first converted back to feature space $\sR^C$ (Latent to Feature, or LtoF in short),
\begin{equation}
    \mathrm{LtoF}(\gZ) = \gX' = \{\rvx'\in\sR^C\}_{i=1,2,\dots, M}.
\end{equation}
The features $\gX'$ are fed into a series self-attention layers to obtain final occupancy/distance function representations $\gF$,
\begin{equation}\label{eq:sa}
    \mathrm{SAs}(\gX') = \gF = \{\rvf\in\sR^C\}_{i=1,2,\dots, M},
\end{equation}
where $\mathrm{SAs}(\cdot)$ is implemented using several self-attention layers. Now we can decode a continuous function. For a continuous coordinate in the space $\sR^3$, we have
\begin{equation}\label{eq:query-func}
    \gO(\mathbf{p}) = \mathrm{FC}\left(\mathrm{CA}(\mathbf{p}, \gF)\right)\in \sR.
\end{equation}
See~\cref{tab:ftl_ltf} for more details on $\mathrm{FtoL}(\cdot)$ and $\mathrm{LtoF}(\cdot)$.

\begin{figure}
    \centering
\scalebox{0.8}{
\begin{tikzpicture}



    \tikzstyle{encoder} = [rectangle, rounded corners, draw=violet!80, fill=violet!10]
    \tikzstyle{decoder} = [rectangle, rounded corners, draw=olive!80, fill=olive!10]
    \tikzstyle{bottleneck} = [rectangle, rounded corners, draw=teal!80, fill=teal!10]

    \pgfdeclarelayer{background}
    \pgfdeclarelayer{box}
    \pgfsetlayers{box, background,main}
    
    \definecolor{mycol1}{RGB}{187,230,191}
    \definecolor{mycol2}{RGB}{170,197,209}
    \definecolor{mycol3}{RGB}{137,230,191}
    \definecolor{mycol4}{RGB}{210,197,209}
    
    \tikzstyle{arrow} = [thick,->,>=stealth]

    \tikzstyle{cross} = [rectangle, draw=mycol1!80, fill=mycol1!50]
    \tikzstyle{self} = [rectangle, draw=mycol2!80, fill=mycol2!50]
    \tikzstyle{ftl} = [rectangle, draw=mycol3!80, fill=mycol3!50]
    \tikzstyle{ltf} = [rectangle, draw=mycol4!80, fill=mycol4!50]


    \node[inner sep=0pt,] (P0) {$\gP_0$};
    \node[right=1cm of P0, inner sep=0pt,] (P01) {$\gP_0$};
    

    \node[cross, below=0.5cm of P01] (CA1) {CA};
    \node[inner sep=0pt,] at (CA1 -| P0) (P1) {$\gP_1$};

    \node[right=0.5cm of CA1, inner sep=0pt,] (X1) {$\gX_1$};

    

    \node[cross, below=0.5cm of X1, ] (CA2) {CA};

    \node[inner sep=0pt,] at (CA2 -| P0) (P2) {$\gP_2$};

    \node[right=0.5cm of CA2, inner sep=0pt,] (X2) {$\gX_2$};



    \node[cross, below=0.5cm of X2] (CA3) {CA};

    \node[inner sep=0pt,] at (CA3 -| P0) (P3) {$\gP_3$};

    \node[right=0.5cm of CA3, inner sep=0pt,] (X3) {$\gX_3$};


    \node[ftl, right=0.5cm of X3] (FTL3) {FtoL};

    \node[right=0.5cm of FTL3] (Z3) {$\gZ_3$};
    
    \node[ltf, right=0.5cm of Z3] (LTF3) {LtoF};

    \node[right=0.5cm of LTF3, inner sep=0pt,] (X31) {$\gX'_3$};
    \node[ftl] at (X2 -| FTL3) (FTL2) {FtoL};

    \node[] at (FTL2 -| Z3) (Z2) {$\gZ_2$};

    \node[ltf, right=0.5cm of Z2] (LTF2) {LtoF};

    \node[right=0.5cm of LTF2, inner sep=0pt,] (X21) {$\gX'_2$};
    \node[ftl]  at (X1 -| FTL2)  (FTL1) {FtoL};

    \node[] at (FTL1 -| Z2)  (Z1) {$\gZ_1$};

    \node[ltf, right=0.5cm of Z1] (LTF1) {LtoF};

    \node[right=0.5cm of LTF1, inner sep=0pt,] (X11) {$\gX'_1$};
    
    \node[self, right=0.5cm of X31] (SA3) {SAs};

    \node[right=0.5cm of SA3, inner sep=0pt,] (F3) {$\gF_3$};

    \node[cross, ] at (X21 -| F3) (upsample2) {CA};

    \node[self, right=0.5cm of upsample2,] (SA2) {SAs};

    \node[right=0.5cm of SA2, inner sep=0pt,] (F2) {$\gF_2$};

    \node[cross, ] at (X11 -| F2) (upsample1) {CA};

    \node[self, right=0.5cm of upsample1,] (SA1) {SAs};

    \node[right=0.5cm of SA1, inner sep=0pt,] (F1) {$\gF_1$};

    \draw [arrow] (P01) -- (CA1);

    \draw [arrow] (P1) -- (CA1);
    \draw [arrow] (CA1) -- (X1);
    
    \draw [arrow] (X1) -- (CA2);
    \draw [arrow] (X1) -- (FTL1);
    \draw [arrow] (FTL1) -- (Z1);
    \draw [arrow] (Z1) -- (LTF1);
    \draw [arrow] (LTF1) -- (X11);

    \draw [arrow] (P2) -- (CA2);
    \draw [arrow] (CA2) -- (X2);

    \draw [arrow] (X2) -- (CA3);
    \draw [arrow] (X2) -- (FTL2);
    \draw [arrow] (FTL2) -- (Z2);
    \draw [arrow] (Z2) -- (LTF2);
    \draw [arrow] (LTF2) -- (X21);

    \draw [arrow] (P3) -- (CA3);
    \draw [arrow] (CA3) -- (X3);

    \draw [arrow] (X3) -- (FTL3);
    \draw [arrow] (FTL3) -- (Z3);
    \draw [arrow] (Z3) -- (LTF3);
    \draw [arrow] (LTF3) -- (X31);
    

    \draw [arrow] (X31) -- (SA3);
    \draw [arrow] (SA3) -- (F3);

    \draw [arrow] (F3) -- (upsample2);
    \draw [arrow] (X21) -- (upsample2);
    \draw [arrow] (upsample2) -- (SA2);
    \draw [arrow] (SA2) -- (F2);

    \draw [arrow] (F2) -- (upsample1);
    \draw [arrow] (X11) -- (upsample1);
    \draw [arrow] (upsample1) -- (SA1);
    \draw [arrow] (SA1) -- (F1);

    \draw [arrow, dashed] (P0) -- (P01);

    \draw [arrow] (P0) to [bend right=30]  node [below, sloped]  () {FPS} (P1);
    \draw [arrow] (P1) to [bend right=30]  node [below, sloped]  () {FPS} (P2);
    \draw [arrow] (P2) to [bend right=30]  node [below, sloped]  () {FPS} (P3);

    \coordinate[] (encode_lower_coord) at (Z3.south -| X3);
    
    \begin{pgfonlayer}{background}
        \node [draw, encoder, fit=(CA1)(X3)(FTL1)(encode_lower_coord), label={below:Encoder}] {}; 
    \end{pgfonlayer}

    \coordinate[] (decode_lower_coord) at (Z3.south -| F3);
    
   \begin{pgfonlayer}{background}
        \node [draw, decoder, fit=(LTF1)(F1)(F3)(decode_lower_coord), label={below:Decoder}] {}; 
    \end{pgfonlayer}

   \begin{pgfonlayer}{background}
        \node [draw, bottleneck, fit=(Z1)(Z2)(Z3), label={below:Bottleneck}] {}; 
    \end{pgfonlayer}
    
\end{tikzpicture}
}
    \vspace{-20pt}
    \caption{\textbf{LaGeM architecture.} We show an illustration with 3 levels of latents.}
    \label{fig:lgm3}
\end{figure}
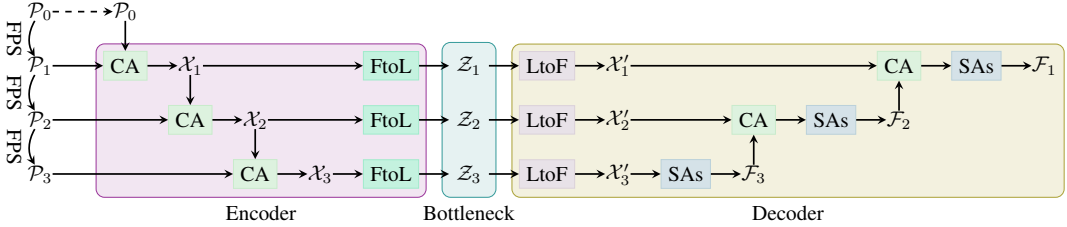

\begin{table}[]
    \centering
    \caption{\textbf{Regularization in the Bottleneck.} We compare the proposed regularization and VAE. We do not need an explicit loss to regularize the latent space.}
    \begin{tabular}{c|cc|c|c}
        \toprule
        & \multicolumn{2}{c|}{Features to Latents (FtoL)} & Latent Loss & Latents to Features (LtoF) \\
        \midrule
        \multirow{2}{*}{VAE} & $\boldsymbol{\mu}=\mathrm{FC}_{\mu}(\rvx)$ & \multirow{2}{*}{$\rvz=\boldsymbol{\mu} +\boldsymbol{\sigma}\odot\boldsymbol{\epsilon}$} & \multirow{2}{*}{KL Divergence}&\multirow{2}{*}{$\rvx' = \mathrm{FC}_{\text{up}}(\rvz)$} \\
        & $\boldsymbol{\sigma}=\mathrm{FC}_{\sigma}(\rvx)$ & & & \\ 
        \rowcolor{mytbcol!30} Ours & $\bar{\rvz}=\mathrm{FC}_{\text{down}}(\rvx)$ & $\displaystyle \rvz=\frac{\bar{\rvz}-\mathrm{E}[\bar{\rvz}]}{\sqrt{\mathrm{Var}[\bar{\rvz}]}}$ & - & $\rvx'=\mathrm{FC}_{\text{up}}(\rvz\odot\boldsymbol{\gamma}+\boldsymbol{\beta})$\\ \bottomrule
    \end{tabular}
    \label{tab:ftl_ltf}
\end{table}
\subsection{Hierarchical VecSet}
The complexity of the self-attention layers in~\cref{eq:sa} is $O(M^2)$, i.e., quadratic in the number of input vectors. This severely affects the training time when $M$ is large. However, to represent high-quality geometry details, we usually need a large $M$. This is making training a large VecSet network more challenging (for example $M=2048$ in CLAY~\citep{clay}). Motivated by the design of the U-Net and NVAE~\citep{vahdat2020nvae}, we propose a new network. Specifically,
in the design of the U-Net (see an illustration in~\cref{fig:ae}), image feature grids are downsampled to lower resolutions where some convolution blocks are applied, and then upsampled to the original resolution. In this way, we can avoid performing convolutional layers in high resolution images (which can be time-consuming). We transferred this idea to the VecSet representations. Two necessary building blocks are operations to down-sample and up-sample a VecSet. Inspired by the design of 3DShape2VecSet~\citep{vecset} (an illustration can be found in~\cref{fig:lgm1}), we interpret the cross attention in the encoder part as a down-sampling operator. Similarly, we can also use it for up-sampling. The resulting network is shown in~\cref{fig:lgm3}.

We have $L$ levels in the U-Net-style transformer, where we number the levels from one (highest resolution) to $L$ (lowest resolution). For notational convenience, we denote the input point cloud as level 0. In the $i$-th level, we first obtain a lower resolution of the point clouds in the $(i-1)$-th level, $\mathrm{FPS}(\gP_{i-1}, r_{i-1})=\gP_{i}$
where $\gP_0$ is the input point cloud. We use cross attention to compress the feature set $\mathrm{CA}(\gP_{i}, \gP_{i-1}) = \gX_{i}$.
Different from previous approaches, we propose a new to way regularize the latent space,
\begin{equation}
    \mathrm{FtoL}(\gX_{i}) = \mathrm{ZeroMeanAndUnitVariance}(\mathrm{FC}_{\text{down}}(\gX_{i})) = \gZ_{i},
\end{equation}
where we normalize each vector in the set to have zero mean and unit variance $(\rvz-\mathrm{E}[\rvz])/\sqrt{\mathrm{Var}[\rvz]}$ (It is often called \emph{standardization} in machine learning which is used to standardize the features present in the data in a fixed range.). The motivation behind this design is that diffusion starts with Gaussian noise which also has zero mean and unit variance. In this way, we enforce both our latent space and the initial Gaussian noise to have similar properties.
To map the latents back to features, we first scale and shift latents back $\rvz\odot\boldsymbol\gamma + \boldsymbol\beta$ (both $\boldsymbol\gamma$ and $\boldsymbol\beta$ are learnable parameters like in Layer Normalization~\citep{lei2016layer}),
\begin{equation}
    \mathrm{LtoF}(\gZ_{i}) = \mathrm{FC}_{\text{up}}(\mathrm{ScaleAndShift}(\gZ_{i}))= \gX'_{i}.
\end{equation}
Unlike KL divergence in a VAE, we do not need an explicit loss term for the latent space. See~\cref{tab:ftl_ltf} for a comparison between the proposed regularization and commonly used KL divergence in VAEs.

\begin{wrapfigure}{c}{0.32\textwidth}
    \begin{center}
    \vspace{-10pt}
        \scalebox{0.7}{

\begin{tikzpicture}[]

    \definecolor{mycol1}{RGB}{187,230,191}
    \definecolor{mycol5}{RGB}{45,197,209}

    \tikzstyle{cross} = [rectangle, draw=mycol1!80, fill=mycol1!50]
    \tikzstyle{fc} = [rectangle, draw=mycol5!80, fill=mycol5!50]

    \tikzstyle{arrow} = [thick,->,>=stealth]

    \node[inner sep=0pt,] (p) {$\rvp$};

    \node[cross, right=1cm of p] (CA2) {CA};
    \node[above=0.5cm of CA2, inner sep=0pt,] (F2) {$\gF_2$};

    \node[cross, above=1.3cm of CA2] (CA1) {CA};
    \node[above=0.5cm of CA1, inner sep=0pt,] (F1) {$\gF_1$};

    \node[cross, below=1.3cm of CA2] (CA3) {CA};
    \node[above=0.5cm of CA3, inner sep=0pt,] (F3) {$\gF_3$};

    \coordinate[right=0.5cm of p] (p_coord);

    \draw [arrow, ] (p) -- (p_coord) |- (CA1);
    \draw [arrow, ] (p) -- (CA2);
    \draw [arrow, ] (p) -- (p_coord) |- (CA3);

    \draw [arrow, ] (F1) -- (CA1);
    \draw [arrow, ] (F2) -- (CA2);
    \draw [arrow, ] (F3) -- (CA3);

    \node[right=0.5cm of CA1, inner sep=0pt,] (f1) {$\rvf_1$};
    \node[right=0.5cm of CA2, inner sep=0pt,] (f2) {$\rvf_2$};
    \node[right=0.5cm of CA3, inner sep=0pt,] (f3) {$\rvf_3$};

    \draw [arrow, ] (CA1) -- (f1);
    \draw [arrow, ] (CA2) -- (f2);
    \draw [arrow, ] (CA3) -- (f3);

    \node [fc, right=1.0cm of f2, ] (FC) {FC};
    \node [right=0.5cm of FC, inner sep=0pt,] (output) {$\gO(\rvp)$};

    \draw [arrow, ] (FC) -- (output);

    \coordinate[left=0.5cm of FC] (fc_coord);

    \draw [arrow, ] (f1) -| (fc_coord) -- (FC);
    \draw [arrow, ] (f2) -- (FC);
    \draw [arrow, ] (f3) -| (fc_coord) -- (FC);
    
\end{tikzpicture}
}
        \vspace{-10pt}
        \caption{Multiresolution Features}
        \label{fig:ca_fc}
    \end{center}
\end{wrapfigure}
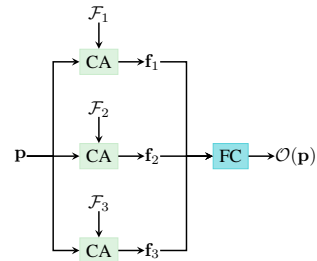

Inspired by the down-sampling usage of cross attention in~\cite{vecset}, we generalize it to \emph{resampling}. Here we use it as \emph{upsampling for unordered set} $\gF_i$.
Before feeding the features to self attention layers, we first upsample features $\gF_{i+1}$ from lower resolution levels and apply self attentions,
\begin{equation}
    \mathrm{SAs}(\mathrm{CA}(\gX'_{i}, \gF_{i+1})) = \gF_{i}.
\end{equation}
The query function in~\cref{eq:query-func} is changed to
\begin{equation}
    \gO(\mathbf{p}) = \mathrm{FC}\left(
        \left[\mathrm{CA}(\mathbf{p}, \gF_{1})| \cdots | \mathrm{CA}(\mathbf{p}, \gF_{L})\right]
    \right)\in\sR,
\end{equation}
where $[\cdot|\cdot|\cdots|\cdot]$ is the symbol for concatenation. This means we are using features from all levels to build the final (occupancy) function representation (\cref{fig:ca_fc}).


\subsection{Diffusion}
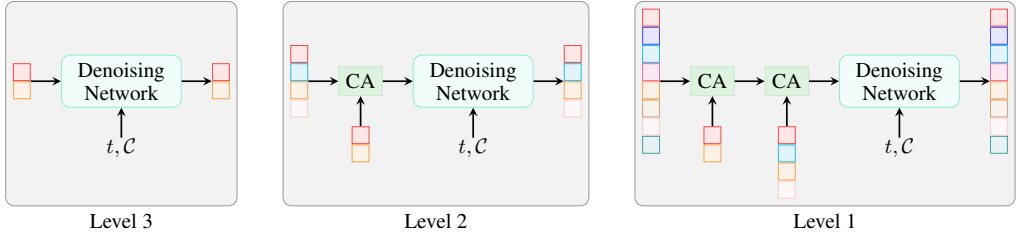
\begin{figure}
    \centering
\scalebox{0.8}{
\begin{tikzpicture}[]

    \tikzset{auto matrix/.style={matrix of nodes,
    inner sep=0pt, ampersand replacement=\&,
    nodes in empty cells,column sep=1.0pt,row sep=0.4pt,
    }}
    
    \pgfsetlayers{background, main}
    \tikzstyle{arrow} = [thick,->,>=stealth]

    \definecolor{mycol1}{RGB}{187,230,191}

    \definecolor{diffcol}{RGB}{115,233,200}
    
    \tikzstyle{diffusion} = [rectangle, rounded corners, draw=diffcol!80, fill=diffcol!10, text width = 5em, text centered]
    \tikzstyle{cross} = [rectangle, draw=mycol1!80, fill=mycol1!50]
    \tikzstyle{stage} = [rectangle, rounded corners, draw=gray!80, fill=gray!10]
    
    \matrix[auto matrix=m,xshift=0em,yshift=0em,opacity=0.9, 
        row 1/.style={nodes={draw=red!80, fill=red!10}},
        row 2/.style={nodes={draw=orange!80, fill=orange!10}},
        cells={nodes={minimum width=0.8em,minimum height=0.8em,
        very thin,anchor=center,
    }}
    ](Z3){
    \\ \\
    };
    \node[diffusion, right=0.5cm of Z3,] (diff3) {Denoising\\Network};

    \node[below=0.5cm of diff3, inner sep=0pt,] (cond3) {$t, \mathcal{C}$};

    \matrix[right=0.5cm of diff3, auto matrix=m,xshift=0em,yshift=0em,opacity=0.9, 
        row 1/.style={nodes={draw=red!80, fill=red!10}},
        row 2/.style={nodes={draw=orange!80, fill=orange!10}},
        cells={nodes={minimum width=0.8em,minimum height=0.8em,
        very thin,anchor=center,
    }}
    ](Z3_out){
    \\ \\
    };

    \draw [arrow] (Z3) -- (diff3);
    \draw [arrow] (diff3) -- (Z3_out);
    \draw [arrow] (cond3) -- (diff3);

    \matrix[right=1.0cm of Z3_out, auto matrix=m,xshift=0em,yshift=0em,opacity=0.9, 
        row 1/.style={nodes={draw=red!80, fill=red!10}},
        row 2/.style={nodes={draw=cyan!80, fill=cyan!10}},
        row 3/.style={nodes={draw=orange!80, fill=orange!10}},
        row 4/.style={nodes={draw=pink!80, fill=pink!10}},
        cells={nodes={minimum width=0.8em,minimum height=0.8em,
        very thin,anchor=center,
    }}
    ](Z2){
    \\ \\ \\ \\
    };

    \node[cross, right=0.5cm of Z2] (CA2) {CA};

    \matrix[below=0.5cm of CA2, auto matrix=m,xshift=0em,yshift=0em,opacity=0.9, 
        row 1/.style={nodes={draw=red!80, fill=red!10}},
        row 2/.style={nodes={draw=orange!80, fill=orange!10}},
        cells={nodes={minimum width=0.8em,minimum height=0.8em,
        very thin,anchor=center,
    }}
    ](Z3_cond){
    \\ \\
    };
    
    \node[diffusion, right=0.5cm of CA2,] (diff2) {Denoising\\Network};

    \node[below=0.5cm of diff2, inner sep=0pt,] (cond2) {$t, \mathcal{C}$};

    \matrix[right=0.5cm of diff2, auto matrix=m,xshift=0em,yshift=0em,opacity=0.9, 
        row 1/.style={nodes={draw=red!80, fill=red!10}},
        row 2/.style={nodes={draw=cyan!80, fill=cyan!10}},
        row 3/.style={nodes={draw=orange!80, fill=orange!10}},
        row 4/.style={nodes={draw=pink!80, fill=pink!10}},
        cells={nodes={minimum width=0.8em,minimum height=0.8em,
        very thin,anchor=center,
    }}
    ](Z2_out){
    \\ \\ \\ \\
    };

    \draw [arrow] (Z2) -- (CA2);
    \draw [arrow] (CA2) -- (diff2);
    \draw [arrow] (diff2) -- (Z2_out);
    \draw [arrow] (cond2) -- (diff2);
    \draw [arrow] (Z3_cond) -- (CA2);

    \matrix[ right=1.0cm of Z2_out, auto matrix=m,xshift=0em,yshift=0em,opacity=0.9, 
        row 1/.style={nodes={draw=red!80, fill=red!10}},
        row 2/.style={nodes={draw=blue!80, fill=blue!10}},
        row 3/.style={nodes={draw=cyan!80, fill=cyan!10}},
        row 4/.style={nodes={draw=magenta!80, fill=magenta!10}},
        row 5/.style={nodes={draw=orange!80, fill=orange!10}},
        row 6/.style={nodes={draw=brown!80, fill=brown!10}},
        row 7/.style={nodes={draw=pink!80, fill=pink!10}},
        row 8/.style={nodes={draw=teal!80, fill=teal!10}},
        cells={nodes={minimum width=0.8em,minimum height=0.8em,
        very thin,anchor=center,
    }}
    ](Z1){
    \\ \\ \\ \\ \\ \\ \\ \\
    };

    \node[cross, right=0.5cm of Z1] (CA1) {CA};

    \matrix[below=0.5cm of CA1, auto matrix=m,xshift=0em,yshift=0em,opacity=0.9, 
        row 1/.style={nodes={draw=red!80, fill=red!10}},
        row 2/.style={nodes={draw=orange!80, fill=orange!10}},
        cells={nodes={minimum width=0.8em,minimum height=0.8em,
        very thin,anchor=center,
    }}
    ](Z3_cond1){
    \\ \\
    };

    \node[cross, right=0.5cm of CA1] (CA11) {CA};
    
    \matrix[below=0.5cm of CA11, auto matrix=m,xshift=0em,yshift=0em,opacity=0.9, 
        row 1/.style={nodes={draw=red!80, fill=red!10}},
        row 2/.style={nodes={draw=cyan!80, fill=cyan!10}},
        row 3/.style={nodes={draw=orange!80, fill=orange!10}},
        row 4/.style={nodes={draw=pink!80, fill=pink!10}},
        cells={nodes={minimum width=0.8em,minimum height=0.8em,
        very thin,anchor=center,
    }}
    ](Z2_cond){
    \\ \\ \\ \\
    };

    \node[diffusion, right=0.5cm of CA11,] (diff1) {Denoising\\Network};

    \node[below=0.5cm of diff1, inner sep=0pt,] (cond1) {$t, \mathcal{C}$};

    \matrix[ right=0.5cm of diff1, auto matrix=m,xshift=0em,yshift=0em,opacity=0.9, 
        row 1/.style={nodes={draw=red!80, fill=red!10}},
        row 2/.style={nodes={draw=blue!80, fill=blue!10}},
        row 3/.style={nodes={draw=cyan!80, fill=cyan!10}},
        row 4/.style={nodes={draw=magenta!80, fill=magenta!10}},
        row 5/.style={nodes={draw=orange!80, fill=orange!10}},
        row 6/.style={nodes={draw=brown!80, fill=brown!10}},
        row 7/.style={nodes={draw=pink!80, fill=pink!10}},
        row 8/.style={nodes={draw=teal!80, fill=teal!10}},
        cells={nodes={minimum width=0.8em,minimum height=0.8em,
        very thin,anchor=center,
    }}
    ](Z1_out){
    \\ \\ \\ \\ \\ \\ \\ \\
    };

    \draw [arrow] (Z1) -- (CA1);
    \draw [arrow] (CA1) -- (CA11);
    \draw [arrow] (CA11) -- (diff1);
    \draw [arrow] (diff1) -- (Z1_out);
    \draw [arrow] (cond1) -- (diff1);
    \draw [arrow] (Z3_cond1) -- (CA1);
    \draw [arrow] (Z2_cond) -- (CA11);

    \coordinate[] (l2_lower) at (Z2_cond.south -| Z3_cond.south);
    \coordinate[] (l2_upper) at (Z1.north -| Z2.north);

    \coordinate[] (l3_lower) at (Z2_cond.south -| cond3.south);
    \coordinate[] (l3_upper) at (Z1.north -| Z3.north);
    
    \begin{pgfonlayer}{background}
        \node [draw, stage, fit=(Z3)(cond3)(Z3_out)(l3_lower)(l3_upper), label={below:Level 3}] {}; 
    \end{pgfonlayer}
    
    \begin{pgfonlayer}{background}
        \node [draw, stage, fit=(Z2)(cond2)(Z3_cond)(Z2_out)(l2_lower)(l2_upper), label={below:Level 2}] {}; 
    \end{pgfonlayer}
    
    \begin{pgfonlayer}{background}
        \node [draw, stage, fit=(Z1)(cond1)(Z2_cond)(Z1_out), label={below:Level 1}] {}; 
    \end{pgfonlayer}

\end{tikzpicture}
}
    \vspace{-10pt}
    \caption{\textbf{Cascaded Latent Diffusion.}}
    \label{fig:diffusion}
\end{figure}

Cascaded Diffusion~\citep{ho2022cascaded} proposed a method for generating high-resolution images. The method is composed of several stages, where each stage is a conditioned diffusion model. Motivated by this, we propose a cascaded latent diffusion model. In Cascaded Diffusion, images generated from the previous stage are used as a condition in the next stage. We build a cascaded latent diffusion model based on Cascaded Diffusion. Formally, the optimization goal (for our three-level implementation) is as follows,
\begin{equation}
    \begin{aligned}
        &\min_{D_3}\left\|D_3(\tilde{\gZ}_3(t), t, \gC\phantom{, \gZ_3, \gZ_2}) - \gZ_3\right\|, \\
        &\min_{D_2}\left\|D_2(\tilde{\gZ}_2(t), t, \gC, \gZ_3\phantom{, \gZ_2}) - \gZ_2\right\|, \\
        &\min_{D_1}\left\|D_1(\tilde{\gZ}_1(t), t, \gC, \gZ_3, \gZ_2) - \gZ_1\right\|, \\
    \end{aligned} 
\end{equation}
where $D_i$ is a denoising network, $t$ represents timestep or noise level, $\tilde{\gZ}_i(t)$ is the noised version (at timestep $t$) of the latent, $\mathcal{C}$ is optional condition information (\textit{e.g.}, text, images, or categories). The network design is based on DiT~\citep{Peebles2022DiT}. To generate latents $\gZ_i$, we need latents from previous stages $\gZ_{>i}$. For diffusion-based image super-resolution methods, this is often done by bilinearly interpolating small images and concatenating them with denoising networks' inputs. As shown in the previous section, we use cross attention for resampling (both down-sampling and upsampling). Here we also utilize cross attention to upsample a latent set. Specifically, assuming we are training a denoising network for $\gZ_2$, the input of the network is $\tilde{\gZ}_2(t)$, 
\begin{equation}
    \mathrm{CA}(\tilde{\gZ}_2(t), \gZ_3).
\end{equation}
Similarly, for $\gZ_1$,
\begin{equation}
    \mathrm{CA}(\mathrm{CA}(\tilde{\gZ}_1(t), \gZ_3), \gZ_2).
\end{equation}
In this way, we are gathering information from previous stages.
See~\cref{fig:diffusion} for an illustration about the pipeline.

\begin{table}[]
    \centering
    \caption{\textbf{Running Statistics of LaGeM.} When using a small number (512) of latent vectors, our model uses 0.87x time and 0.66x memory during training. For larger models (2k latent vectors), the advantage is even more significant (0.7x time and 0.58x memory).}
\def\arraystretch{1.15}\tabcolsep=0.47em
\begin{tabular}{c|cc|cc|cc}
\toprule
              & VecSet & LaGeM    & VecSet & LaGeM       & VecSet & LaGeM       \\ \midrule
\rowcolor{mytbcol!30}Batch Size    & \multicolumn{2}{c|}{64}     &   \multicolumn{2}{c|}{8} & \multicolumn{2}{c}{4}    \\
Self Attn Layers   & 24     & 8/8/8       & 24     & 8/8/8          & 24     & 8/8/8          \\
\rowcolor{mytbcol!30}Attn Channels & 512    & 512/512/512 & 1k   & 1k/1k/1k & 1k   & 1k/1k/1k \\
\# Parameters (M) & 106.13 & 125.15      & 424.24 & 499.85         & 424.24 & 499.85         \\ \midrule
\rowcolor{mytbcol!30}\# Latent Vectors        & 512    & 32/128/512  & 2k   & 128/512/2k   & 2k   & 128/512/2k   \\ 
\# Latent Channels & 8 & 32/16/8 & 64 & 64/32/16 & 64 & 64/32/16 \\
\midrule
\rowcolor{mytbcol!30} Training Memory (M)       & 56,125  & \textbf{37,055} \underline{\scriptsize{(0.66$\times$)}}       & OOM    & \textbf{53,791} \underline{\scriptsize{(-)}}         & 54,543  & \textbf{31,662} \underline{\scriptsize{(0.58$\times$)}}          \\
Training Iteration (sec)        & 0.6481 & \textbf{0.5658} \underline{\scriptsize{(0.87$\times$)}}      & -      & \textbf{0.7714} \underline{\scriptsize{(-)}}         & 0.6902 & \textbf{0.4853} \underline{\scriptsize{(0.70$\times$)}}         \\
\bottomrule
\end{tabular}
    \label{tab:stat}
\end{table}
\section{Experiments}
\subsection{Autoencoding Model}

\begin{figure}
    \centering
    \includegraphics[width=\linewidth]{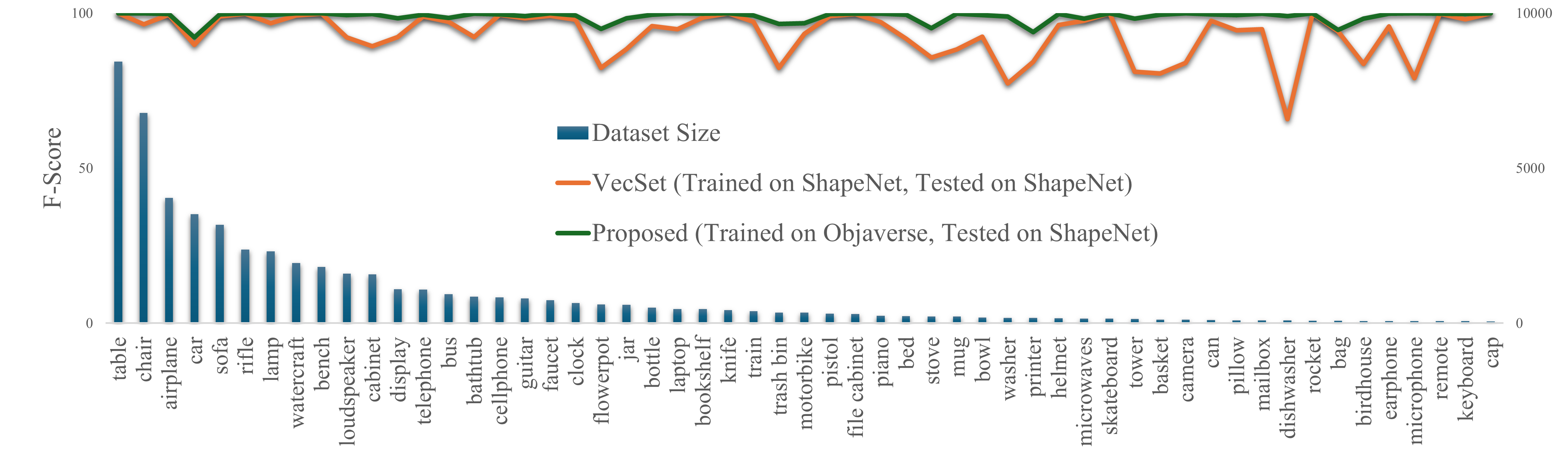}
    \caption{\textbf{Generalization on ShapeNet.} Our results are better than VecSet in all categories. On small categories, the results of VecSet are not stable because of limited training samples. In contrast, our trained model also performs well in these categories.}
    \label{fig:shapenet-all}
\end{figure}

The main autoencoding experiment is trained on Objaverse~\citep{deitke2023objaverse}. Models are zero-centered and normalized into the unit sphere. Since most 3D models in this dataset are not watertight, we use ManifoldPlus~\citep{huang2020manifoldplus} to make all meshes watertight. Due to failures of modeling loading and conversion, we obtained around 600k watertight models for training. The three levels of latents are $128\times 64$, $512\times 32$, and $2048\times 16$ (where 64, 32, and 16 are channels of the latents). Some other hyperparameters of the network can also be found in~\cref{tab:stat}. We name the model as LaGeM-Objaverse. We also apply the method to ShapeNet, where the train split is taken from~\citep{zhang20223dilg}. Since ShapeNet is a relatively small and easy dataset compared to Objaverse, we choose smaller latents which are $32\times 32$, $128\times 16$, and $512\times 8$. The model is named as LaGeM-ShapeNet. Both models are compared against VecSet~\citep{vecset}. We use Chamfer distance and F-score as the metrics. The results are shown in~\cref{tab:main}. Like~\citep{vecset}, we first compare the results on the largest categories (which have several thousand training samples) in ShapeNet and then all categories. We can see that, LaGeM-ShapeNet has almost the same number of parameters as VecSet, but with much shorter training time and less training memory. The quantitative results (averaged over all ShapeNet categories) are also better than VecSet's. While for LaGeM-Objaverse, there is a large improvement in both training cost and quantitative results. The quantitative results show an improvement of almost 50 percent averaged across the complete dataset in terms of the metric Chamfer. This demonstrates that LaGeM-Objaverse has good generalization ability. This can also be seen in~\cref{fig:shapenet-all}. The results of LaGeM-Objaverse are good on small categories of ShapeNet. In previous works~\citep{vecset}, this is nearly impossible because of limited training samples.
\begin{table}[]
    \centering
    \caption{\textbf{Evaluation on ShapeNet.} We compare our results to VecSet~\citep{vecset} trained on ShapeNet. If we train our model on ShapeNet and evaluate on ShapeNet our model is slightly better than VecSet. When our model is trained on Objaverse and evaluated on ShapeNet, we can see a very large improvement. Note that it is difficult to scale VecSet to Objaverse training.}
\begin{tabular}{r|>{\columncolor{gray!10}}r>{\columncolor{mytbcol!30}}rr>{\columncolor{mytbcol!30}}rr|>{\columncolor{gray!10}}r>{\columncolor{mytbcol!30}}rr>{\columncolor{mytbcol!30}}rr}
\toprule
         &\multicolumn{5}{c|}{Chamfer $\downarrow (\times 100)$}& \multicolumn{5}{c}{F-Score $\uparrow (\times 100)$ } \\
         &\multicolumn{1}{c}{\multirow{2}{*}{VS}}& \multicolumn{4}{c|}{LaGeM($\Delta$)}& \multicolumn{1}{c}{\multirow{2}{*}{VS}} & \multicolumn{4}{c}{LaGeM($\Delta$)}\\

         &\multicolumn{1}{c}{}&      \multicolumn{2}{c}{ShapeNet}   &     \multicolumn{2}{c|}{Objaverse}  & \multicolumn{1}{c}{}&      \multicolumn{2}{c}{ShapeNet}   &     \multicolumn{2}{c}{Objaverse} \\ \midrule
table    & 2.46 & 2.48 & 0.02  & \textbf{2.09} & -0.37 & 99.94 & 99.97 & 0.02  & \textbf{99.96} & 0.02 \\
car      & 5.99 & 5.89 & -0.10 & \textbf{4.36} & -1.63 & 89.85 & 90.31 & 0.46  & \textbf{92.15} & 2.30 \\
chair    & 2.92 & 2.89 & -0.03 & \textbf{2.01} & -0.91 & 96.40 & 96.49 & 0.09  & \textbf{99.91} & 3.51 \\
airplane & 1.78 & 1.81 & 0.03  & \textbf{1.58} & -0.21 & 99.50 & 99.48 & -0.02 & \textbf{99.78} & 0.29 \\
sofa     & 2.64 & 2.63 & -0.01 & \textbf{2.25} & -0.39 & 98.92 & 99.04 & 0.11  & \textbf{99.60} & 0.67 \\
rifle    & 1.78 & 1.77 & -0.01 & \textbf{1.44} & -0.34 & 99.88 & 99.88 & -0.01 & \textbf{99.94} & 0.06 \\
lamp     & 4.36 & 4.44 & 0.08  & \textbf{2.37} & -2.00 & 96.78 & 97.18 & 0.39  & \textbf{99.43} & 2.64 \\ \midrule
mean (selected) & 3.13 & 3.13 & 0.00  & \textbf{2.30} & -0.83 & 97.33 & 97.48 & 0.15  & \textbf{98.68} & 1.36 \\
mean (all) & 4.68 & 4.63 & -0.04 & \textbf{2.42} & -2.26 & 93.25 & 93.47 & 0.23  & \textbf{98.93} & 5.68 \\
\bottomrule
\end{tabular}

    \label{tab:main}
\end{table}

\begin{table}[]
    \centering
    \caption{\textbf{Generalization on Various Datasets.} Our trained model is capable of doing inference on several existing datasets. It can be applied on non-watertight datasets like ABO and pix3d even the model is trained on watertight datasets. Note that models from ShapeNet are not watertight originally. We use the watertight version processed by~\citep{zhang20223dilg}. The metric for ShapeNet-test is different from~\cref{tab:main}. It is because here we show metrics averaged over all objects instead of categories.}
\def\arraystretch{1.15}\tabcolsep=0.32em
\begin{tabular}{ccc|>{\columncolor{gray!10}}r>{\columncolor{mytbcol!30}}rr|>{\columncolor{gray!10}}r>{\columncolor{mytbcol!30}}rr}
\toprule
&           & & \multicolumn{3}{c|}{Chamfer $\downarrow (\times 100)$} & \multicolumn{3}{c}{F-Score $\uparrow (\times 100)$ } \\ 
\multirow{-2}{*}{Dataset} & \multirow{-2}{*}{\# Meshes} & \multirow{-2}{*}{Manifold} & \multicolumn{1}{c}{VS} & \multicolumn{2}{c|}{LaGeM($\Delta$)} & \multicolumn{1}{c}{VS} 
 & \multicolumn{2}{c}{LaGeM($\Delta$)}\\ \midrule
Thingi10k~\citep{zhou2016thingi10k} & 10k & Yes & 4.52 & \textbf{2.99} & -1.53 & 92.75 & \textbf{97.19} & 4.44\\
ABO~\citep{collins2022abo} & 8k & No & 4.91	&\textbf{3.66}	&-1.26	&92.52	&\textbf{94.91}	&2.39\\
ShapeNet~\citep{chang2015shapenet}-test & 2k & Yes & 3.25	& \textbf{2.33}	& -0.92	& 97.41	& \textbf{99.49}	& 2.08 \\
EGAD~\citep{morrison2020egad} & 2k & Yes & 3.27	& \textbf{2.82}	& -0.45	& 99.02	&\textbf{99.76}&	0.74\\
GSO~\citep{downs2022google} & 1k & Yes & 3.78	& \textbf{2.35}	& -1.43	& 94.70 &	\textbf{99.54}	& 4.84\\
pix3d~\citep{sun2018pix3d} & 700 & No & 6.53	& \textbf{6.02}	& -0.50	& 87.25	& \textbf{87.96}	& 0.71\\
FAUST~\citep{bogo2014faust} & 100 & Yes & 2.10 &	\textbf{1.31}	& -0.79	& 99.58	& \textbf{99.90} &	0.32\\
\bottomrule
\end{tabular}
    \label{tab:generalization}
\end{table}

\begin{figure}
    \centering
    \begin{overpic}[trim={0cm 0cm 0cm 0cm},clip,
    width=0.95\linewidth,
    grid=false]{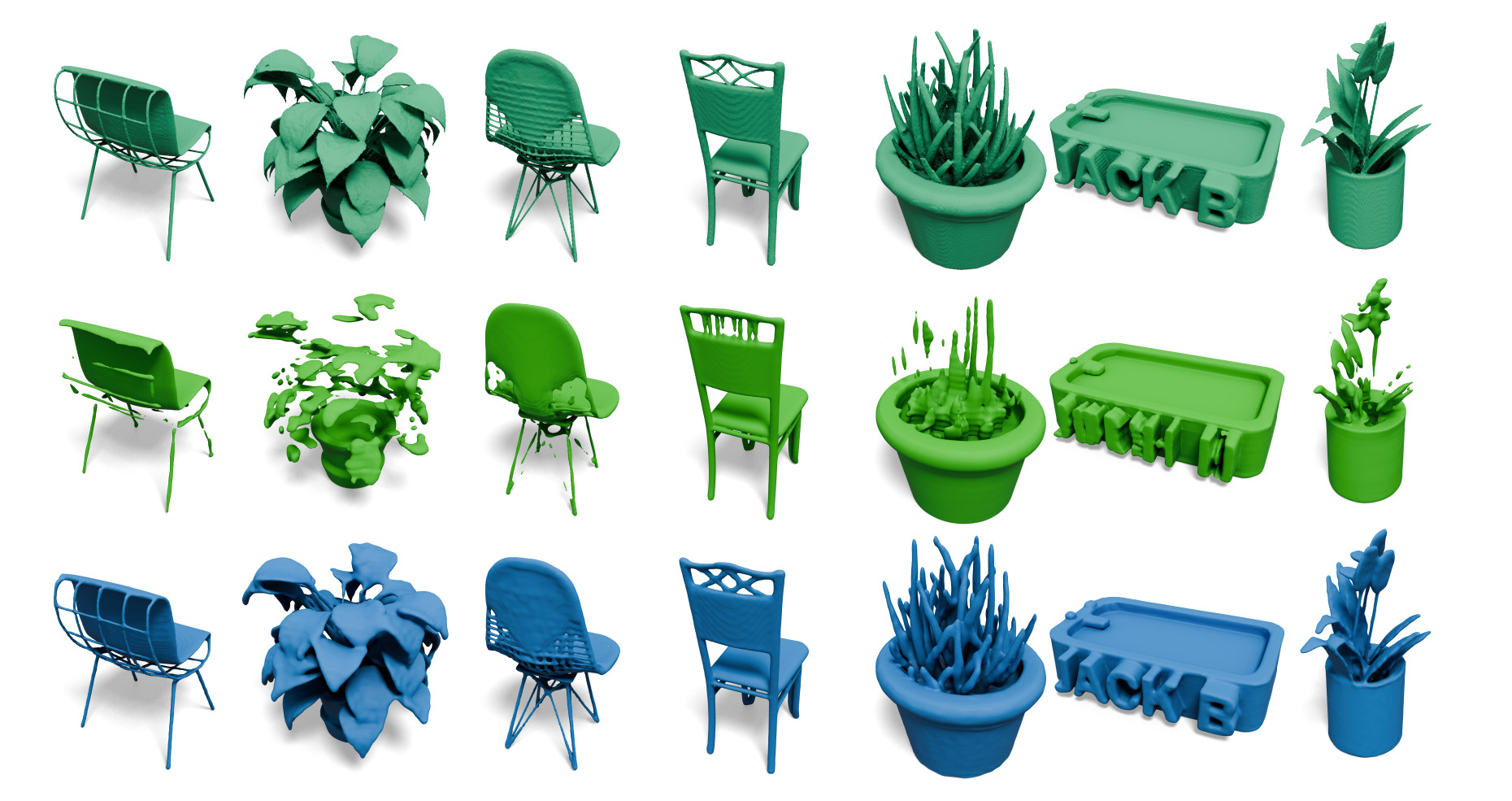}
        \put(-2,4){\rotatebox{90}{LaGeM}}
        \put(-2,22){\rotatebox{90}{VecSet}}
        \put(-2,42){\rotatebox{90}{GT}}
    \end{overpic}
    \vspace{-10pt}
    \caption{\textbf{Qualitative Results on ShapeNet.} We show autoencoding results on ShapeNet. We use VecSet as the baseline. Our model is capable of reconstructing detailed geometry, especially thin structures.}
    \label{fig:shapenet}
\end{figure}

\begin{figure}
    \centering
    \begin{overpic}[trim={0cm 0cm 0cm 0cm},clip,
    width=0.95\linewidth,
    grid=false]{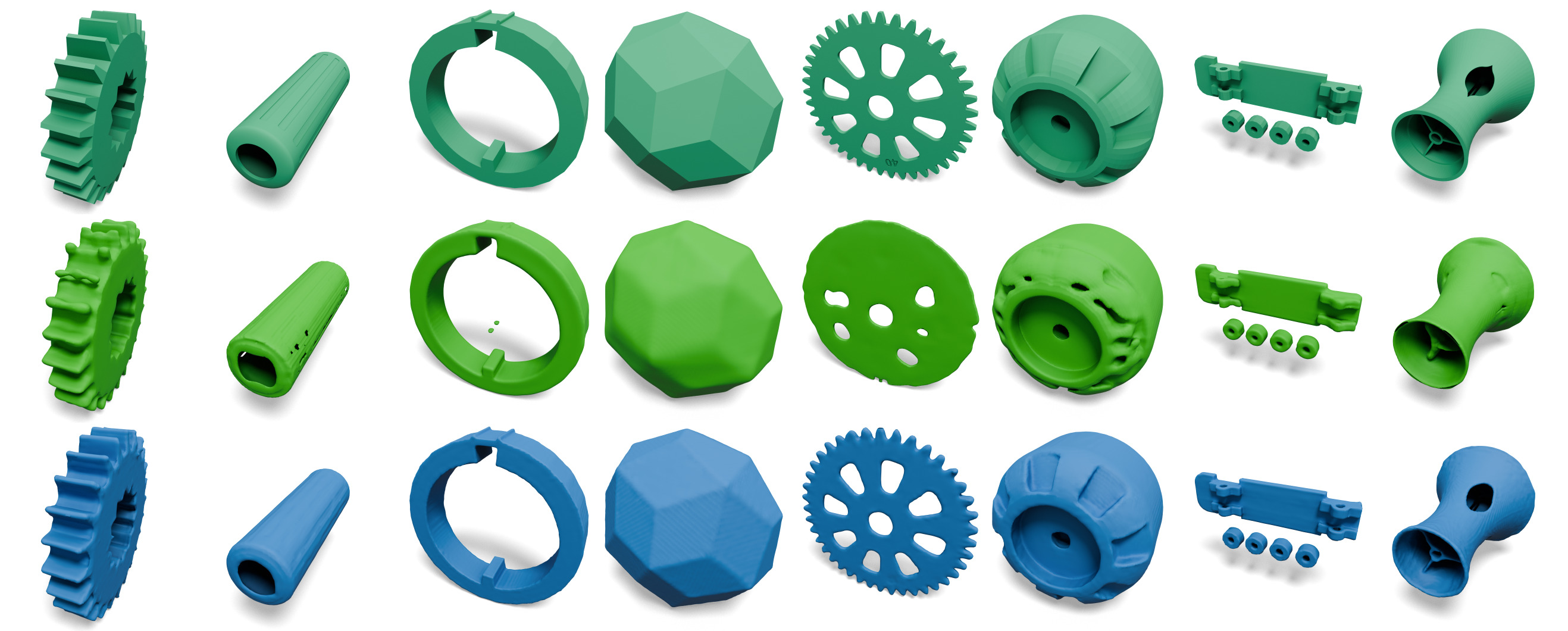}
        \put(-2,3){\rotatebox{90}{LaGeM}}
        \put(-2,16){\rotatebox{90}{VecSet}}
        \put(-2,32){\rotatebox{90}{GT}}
    \end{overpic}
    \vspace{-10pt}
    \caption{\textbf{Qualitative Results on Thingi10k.} Our model can even preserve highly detailed geometry in CAD models.}
    \label{fig:thingi10k}
\end{figure}

\begin{figure}
    \centering
    \begin{overpic}[trim={0cm 0cm 0cm 0cm},clip,
    width=0.95\linewidth,
    grid=false]{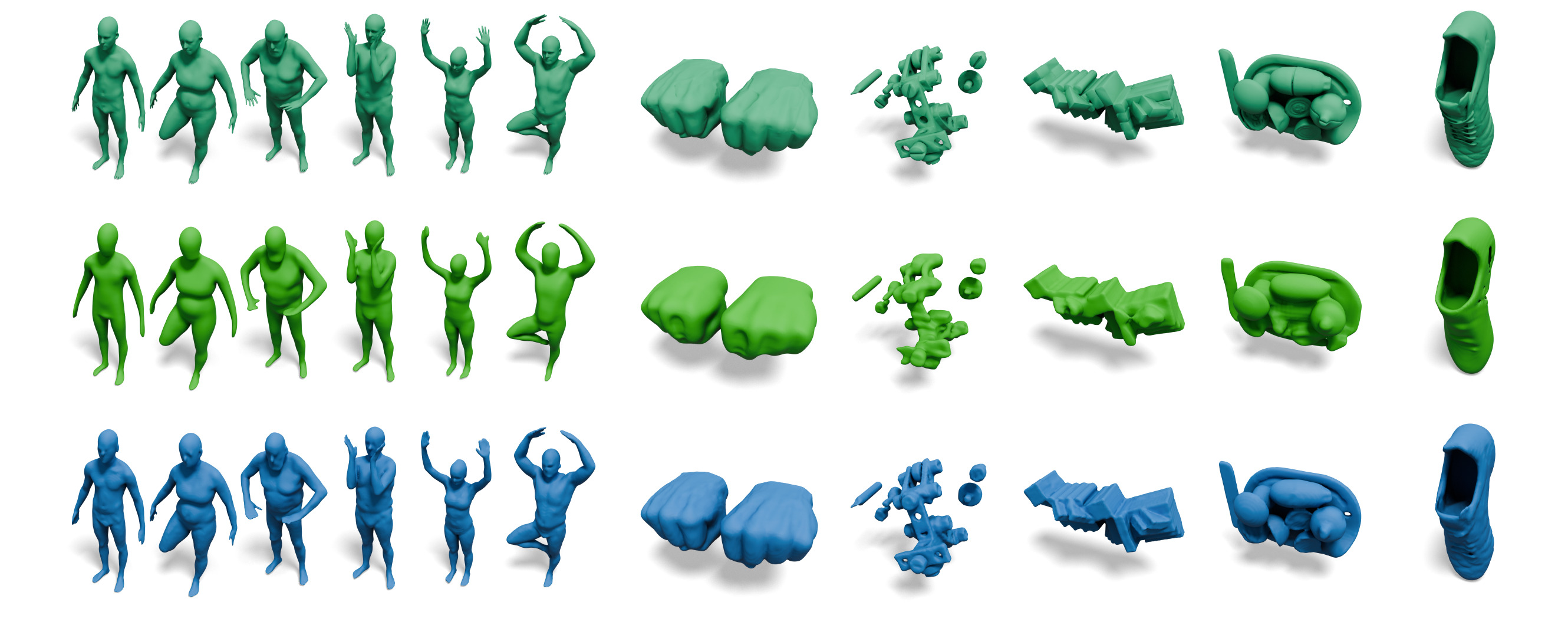}
        \put(-2,3){\rotatebox{90}{LaGeM}}
        \put(-2,16){\rotatebox{90}{VecSet}}
        \put(-2,32){\rotatebox{90}{GT}}
        \put(18,40){FAUST}
        \put(68,40){GSO}
        \dashline{0.7}(39,2)(39,38)
    \end{overpic}
    \vspace{-15pt}
    \caption{\textbf{Qualitative Results on FAUST and GSO.} Results of VecSet are over-smoothed, while our method can preserve sharp details.}
    \label{fig:dataset}
\end{figure}
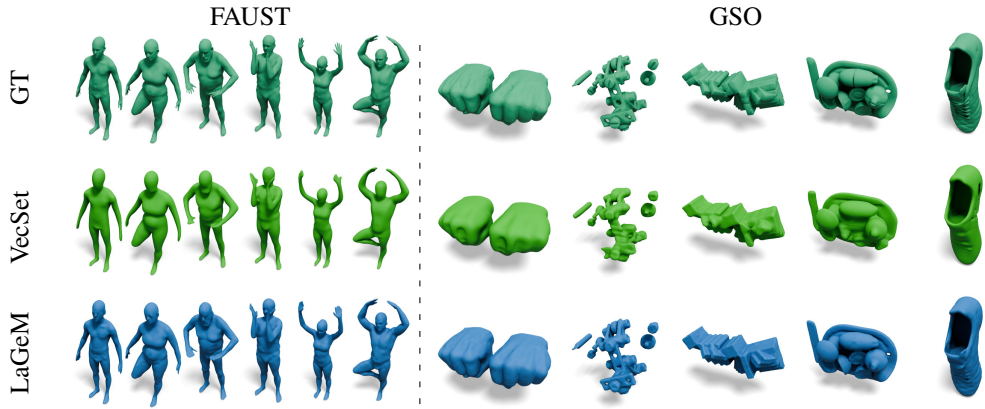

To further prove the generalization ability of LaGeM-Objaverse, we also test the autoencoding on various datasets, including Thingi10k~\citep{zhou2016thingi10k}, ABO~\citep{collins2022abo}, EGAD~\citep{morrison2020egad}, GSO~\citep{downs2022google}, pix3d~\citep{sun2018pix3d} and FAUST~\citep{bogo2014faust}. The objects from these datasets vary from daily objects, CAD models, human models, and synthetic objects. The quantitative results can be found in~\cref{tab:generalization}. We again use VecSet's model as the baseline. From the metrics, we can see that LaGeM-Objaverse is able to represent different kinds of objects with highly detailed geometry and sharp features. Note that, even for non-watertight meshes, the model is still able to do reconstruction. Visual results of the method can be found in~\cref{fig:shapenet}, \cref{fig:thingi10k}, \cref{fig:dataset}.

\subsection{Generative Model}
\begin{figure}
    \centering
    \includegraphics[trim={2cm, 1cm, 2cm, 1cm}, width=1.0\linewidth]{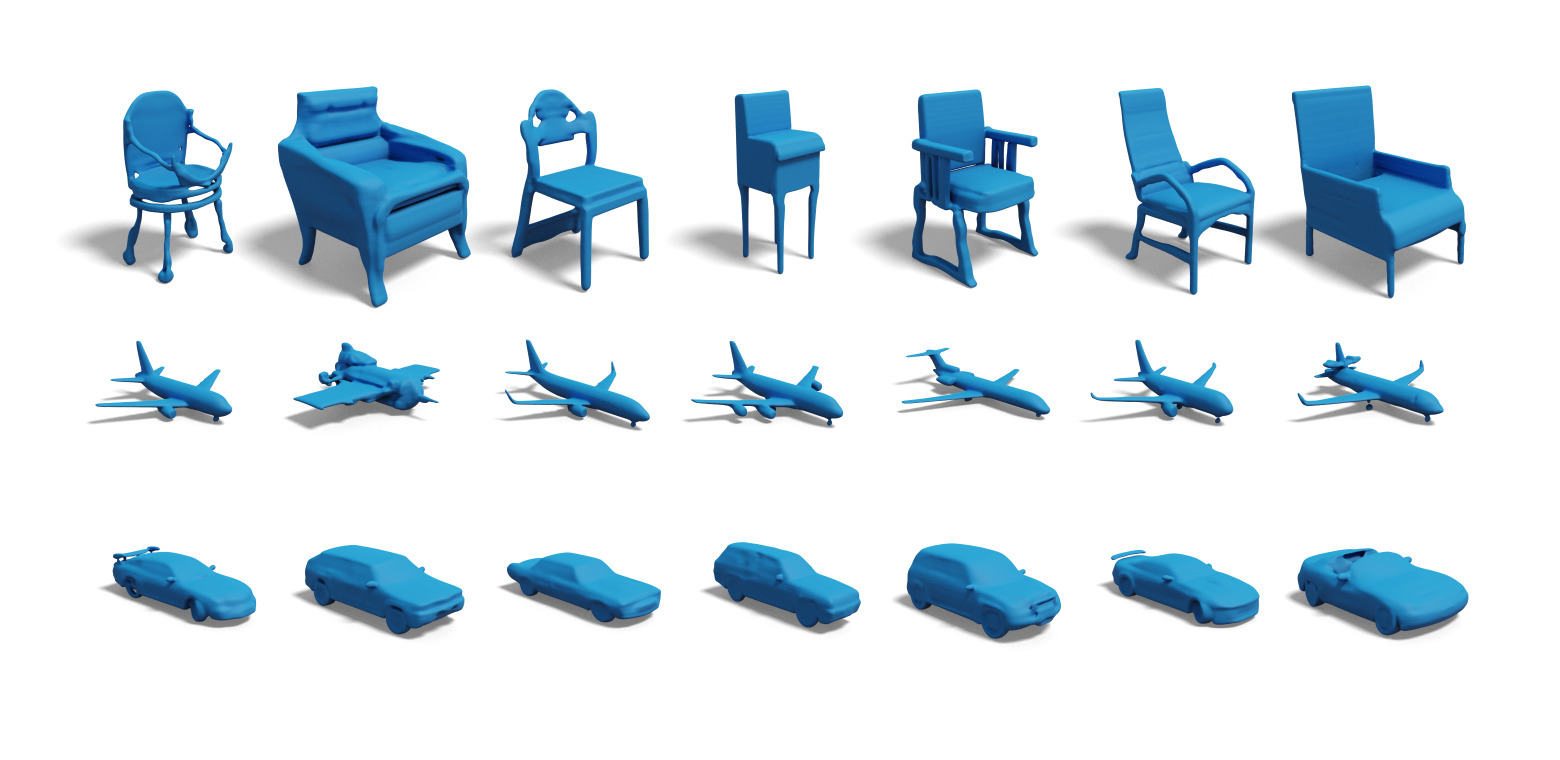}
    \vspace{-30pt}
    \caption{\textbf{Category-Conditioned Generative Results on ShapeNet.}}
    \label{fig:shapenet_gen}
\end{figure}

\begin{figure}
    \centering
    \includegraphics[trim={1cm, 1cm, 1cm, 0cm}, width=1.0\linewidth]{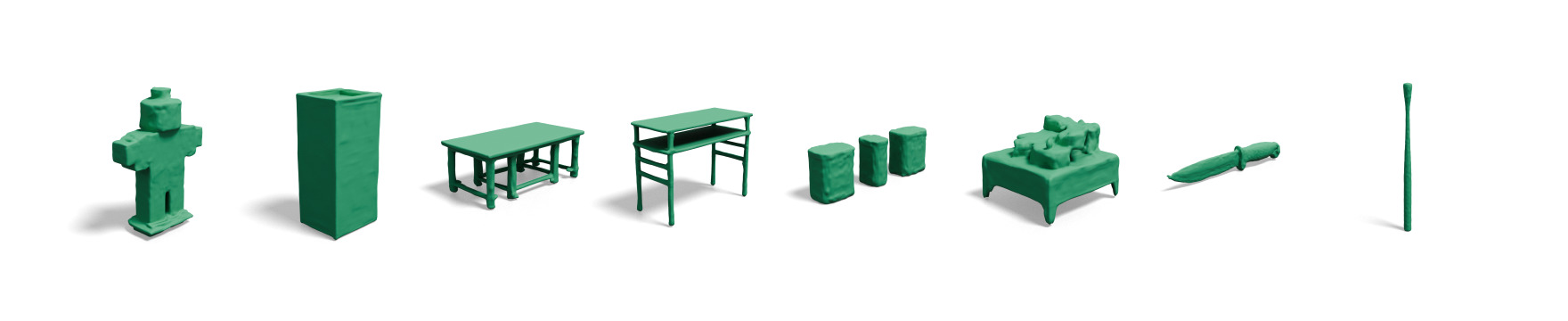}
    \vspace{-20pt}
    \caption{\textbf{Unconditional Generative Results on Objaverse-10k.}}
    \label{fig:objaverse_gen}
\end{figure}

We conducted two generative experiments, one is on ShapeNet with categories as the condition, and the other one is unconditional generation on Objaverse-10k.
For ShapeNet, the denoising networks of the 3 levels have 12 self-attention blocks with 768 channels. We trained the model for around 200 hours with 4 A100 GPUs. The results are shown in~\cref{fig:shapenet_gen}.
For Objaverse-10k, due to limited training GPU resources, we select a subset of 10k models from Objaverse and train the unconditional generative model. There are 24 self-attention blocks with 768 channels in all stages of the latents. The model is trained on 16 A100 GPUs for around 100 hours. See~\cref{fig:objaverse_gen} for some unconditional generation results.

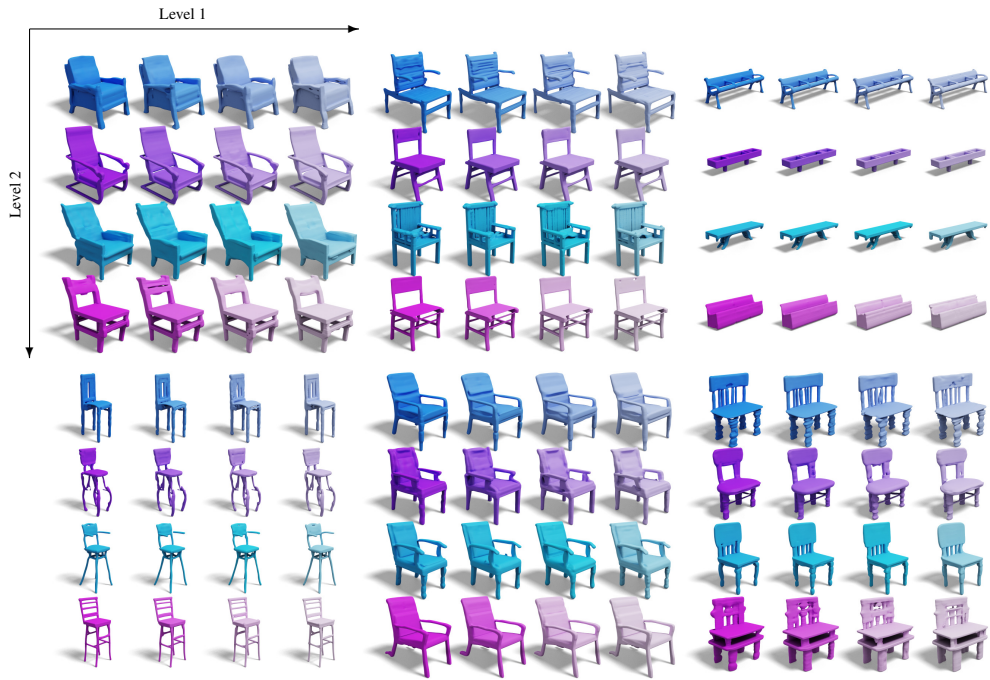
\begin{figure}
    \centering
    \begin{overpic}[trim={0cm 0cm 0cm 0cm},clip,
    width=0.95\linewidth,
    grid=false]{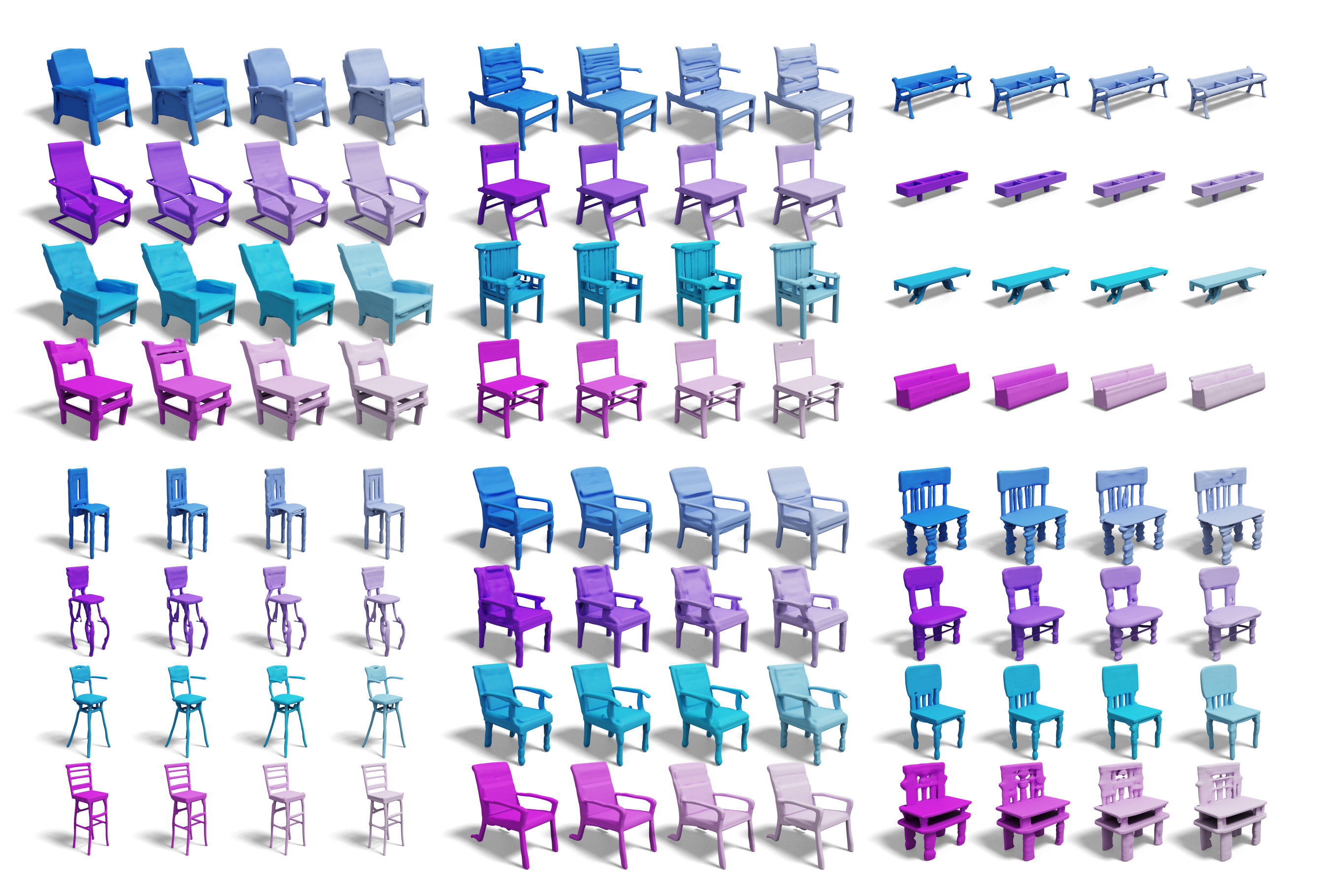}
        \put(13,68){\tiny{Level 1}}
        \put(-2,48){\rotatebox{90}{\tiny{Level 2}}}
        \put(0,67){\vector(0,-1){33}}
        \put(0,67){\vector(1,0){33}}
    \end{overpic}
    \vspace{-10pt}
    \caption{\textbf{Latent Levels.} Each small $4\times 4$ block shares the same level 3 latents $\gZ_3$. 3D models in the same block have similar structures. In each block, every $1\times 4$ line shares the same level 2 latents $\gZ_2$. In each line of a block, 3D models look almost the same except for some minor details. Thus, we argue that $\gZ_3$ controls the \emph{structure}, $\gZ_2$ affects the \emph{major details} and $\gZ_1$ is responsible for \emph{minor details}.}
    \label{fig:levels}
\end{figure}

\paragraph{Controllability of the Latents.} We verify that different levels of latents control different levels of detail of the generated samples. During generation, we first generate higher-level latents $\gZ_3$, which determine the main structures of the 3D models. Then we use $\gZ_3$ as a condition to generate $\gZ_2$, which adds major details to the models. In the end, we generate $\gZ_1$ conditioned on both $\gZ_3$ and $\gZ_2$. This final step adds some minor details to the samples. A visual illustration can be found in~\cref{fig:levels}.


\section{Conclusion}
We proposed LaGeM (Large Geometry Model), an architecture for encoding 3D geometry. Different from previous approaches, the latent space is modeled as a hierarchical latent VecSets. To make this work, our model employs a U-Net-style design and a new regularization technique for the bottleneck. We showed that this model can be trained much faster with much lower GPU memory costs, especially for larger networks and datasets. This enables scaling of the network for large-scale datasets. We release our model trained on a 600k geometry dataset. Additionally, we proposed a cascaded diffusion model to show some preliminary generative results with the hierarchical latent space.


\paragraph{Limitation.} Since the latent space is divided into multiple levels, training a diffusion model on all levels still takes a lot of time. Our method does not solve the high training cost problem of diffusion itself.


\clearpage
\bibliography{bib}

\begin{thebibliography}{55}
\providecommand{\natexlab}[1]{#1}
\providecommand{\url}[1]{\texttt{#1}}
\expandafter\ifx\csname urlstyle\endcsname\relax
  \providecommand{\doi}[1]{doi: #1}\else
  \providecommand{\doi}{doi: \begingroup \urlstyle{rm}\Url}\fi

\bibitem[Bogo et~al.(2014)Bogo, Romero, Loper, and Black]{bogo2014faust}
Federica Bogo, Javier Romero, Matthew Loper, and Michael~J Black.
\newblock Faust: Dataset and evaluation for 3d mesh registration.
\newblock In \emph{Proceedings of the IEEE conference on computer vision and pattern recognition}, pp.\  3794--3801, 2014.

\bibitem[Cao et~al.(2024)Cao, Luo, Zhang, Nie{\ss}ner, and Tang]{cao2024motion2vecsets}
Wei Cao, Chang Luo, Biao Zhang, Matthias Nie{\ss}ner, and Jiapeng Tang.
\newblock Motion2vecsets: 4d latent vector set diffusion for non-rigid shape reconstruction and tracking.
\newblock In \emph{Proceedings of the IEEE/CVF Conference on Computer Vision and Pattern Recognition}, pp.\  20496--20506, 2024.

\bibitem[Chang et~al.(2015)Chang, Funkhouser, Guibas, Hanrahan, Huang, Li, Savarese, Savva, Song, Su, et~al.]{chang2015shapenet}
Angel~X Chang, Thomas Funkhouser, Leonidas Guibas, Pat Hanrahan, Qixing Huang, Zimo Li, Silvio Savarese, Manolis Savva, Shuran Song, Hao Su, et~al.
\newblock Shapenet: An information-rich 3d model repository.
\newblock \emph{arXiv preprint arXiv:1512.03012}, 2015.

\bibitem[Chen et~al.(2023)Chen, Chen, Jiao, and Jia]{chen2023fantasia3d}
Rui Chen, Yongwei Chen, Ningxin Jiao, and Kui Jia.
\newblock Fantasia3d: Disentangling geometry and appearance for high-quality text-to-3d content creation.
\newblock In \emph{Proceedings of the IEEE/CVF international conference on computer vision}, pp.\  22246--22256, 2023.

\bibitem[Chen et~al.(2024{\natexlab{a}})Chen, Chen, Pang, Zeng, Cheng, Fu, Yin, Wang, Wang, Zhang, et~al.]{chen2024meshxl}
Sijin Chen, Xin Chen, Anqi Pang, Xianfang Zeng, Wei Cheng, Yijun Fu, Fukun Yin, Yanru Wang, Zhibin Wang, Chi Zhang, et~al.
\newblock Meshxl: Neural coordinate field for generative 3d foundation models.
\newblock \emph{arXiv preprint arXiv:2405.20853}, 2024{\natexlab{a}}.

\bibitem[Chen et~al.(2024{\natexlab{b}})Chen, He, Huang, Ye, Chen, Tang, Chen, Cai, Yang, Yu, et~al.]{chen2024meshanything}
Yiwen Chen, Tong He, Di~Huang, Weicai Ye, Sijin Chen, Jiaxiang Tang, Xin Chen, Zhongang Cai, Lei Yang, Gang Yu, et~al.
\newblock Meshanything: Artist-created mesh generation with autoregressive transformers.
\newblock \emph{arXiv preprint arXiv:2406.10163}, 2024{\natexlab{b}}.

\bibitem[Chen et~al.(2024{\natexlab{c}})Chen, Tang, Dong, Cao, Hong, Lan, Wang, Xie, Wu, Saito, et~al.]{chen20243dtopia}
Zhaoxi Chen, Jiaxiang Tang, Yuhao Dong, Ziang Cao, Fangzhou Hong, Yushi Lan, Tengfei Wang, Haozhe Xie, Tong Wu, Shunsuke Saito, et~al.
\newblock 3dtopia-xl: Scaling high-quality 3d asset generation via primitive diffusion.
\newblock \emph{arXiv preprint arXiv:2409.12957}, 2024{\natexlab{c}}.

\bibitem[Cheng et~al.(2023)Cheng, Lee, Tulyakov, Schwing, and Gui]{cheng2023sdfusion}
Yen-Chi Cheng, Hsin-Ying Lee, Sergey Tulyakov, Alexander~G Schwing, and Liang-Yan Gui.
\newblock Sdfusion: Multimodal 3d shape completion, reconstruction, and generation.
\newblock In \emph{Proceedings of the IEEE/CVF Conference on Computer Vision and Pattern Recognition}, pp.\  4456--4465, 2023.

\bibitem[Collins et~al.(2022)Collins, Goel, Deng, Luthra, Xu, Gundogdu, Zhang, Vicente, Dideriksen, Arora, et~al.]{collins2022abo}
Jasmine Collins, Shubham Goel, Kenan Deng, Achleshwar Luthra, Leon Xu, Erhan Gundogdu, Xi~Zhang, Tomas F~Yago Vicente, Thomas Dideriksen, Himanshu Arora, et~al.
\newblock Abo: Dataset and benchmarks for real-world 3d object understanding.
\newblock In \emph{Proceedings of the IEEE/CVF conference on computer vision and pattern recognition}, pp.\  21126--21136, 2022.

\bibitem[Deitke et~al.(2023)Deitke, Schwenk, Salvador, Weihs, Michel, VanderBilt, Schmidt, Ehsani, Kembhavi, and Farhadi]{deitke2023objaverse}
Matt Deitke, Dustin Schwenk, Jordi Salvador, Luca Weihs, Oscar Michel, Eli VanderBilt, Ludwig Schmidt, Kiana Ehsani, Aniruddha Kembhavi, and Ali Farhadi.
\newblock Objaverse: A universe of annotated 3d objects.
\newblock In \emph{Proceedings of the IEEE/CVF Conference on Computer Vision and Pattern Recognition}, pp.\  13142--13153, 2023.

\bibitem[Dong et~al.(2024)Dong, Zuo, Gu, Yuan, Zhao, Dong, Bo, and Huang]{dong2024gpld3d}
Yuan Dong, Qi~Zuo, Xiaodong Gu, Weihao Yuan, Zhengyi Zhao, Zilong Dong, Liefeng Bo, and Qixing Huang.
\newblock Gpld3d: Latent diffusion of 3d shape generative models by enforcing geometric and physical priors.
\newblock In \emph{Proceedings of the IEEE/CVF Conference on Computer Vision and Pattern Recognition}, pp.\  56--66, 2024.

\bibitem[Downs et~al.(2022)Downs, Francis, Koenig, Kinman, Hickman, Reymann, McHugh, and Vanhoucke]{downs2022google}
Laura Downs, Anthony Francis, Nate Koenig, Brandon Kinman, Ryan Hickman, Krista Reymann, Thomas~B McHugh, and Vincent Vanhoucke.
\newblock Google scanned objects: A high-quality dataset of 3d scanned household items.
\newblock In \emph{2022 International Conference on Robotics and Automation (ICRA)}, pp.\  2553--2560. IEEE, 2022.

\bibitem[Erko{\c{c}} et~al.(2023)Erko{\c{c}}, Ma, Shan, Nie{\ss}ner, and Dai]{erkocc2023hyperdiffusion}
Ziya Erko{\c{c}}, Fangchang Ma, Qi~Shan, Matthias Nie{\ss}ner, and Angela Dai.
\newblock Hyperdiffusion: Generating implicit neural fields with weight-space diffusion.
\newblock In \emph{Proceedings of the IEEE/CVF international conference on computer vision}, pp.\  14300--14310, 2023.

\bibitem[Ho et~al.(2022)Ho, Saharia, Chan, Fleet, Norouzi, and Salimans]{ho2022cascaded}
Jonathan Ho, Chitwan Saharia, William Chan, David~J Fleet, Mohammad Norouzi, and Tim Salimans.
\newblock Cascaded diffusion models for high fidelity image generation.
\newblock \emph{Journal of Machine Learning Research}, 23\penalty0 (47):\penalty0 1--33, 2022.

\bibitem[Huang et~al.(2020)Huang, Zhou, and Guibas]{huang2020manifoldplus}
Jingwei Huang, Yichao Zhou, and Leonidas Guibas.
\newblock Manifoldplus: A robust and scalable watertight manifold surface generation method for triangle soups.
\newblock \emph{arXiv preprint arXiv:2005.11621}, 2020.

\bibitem[Hui et~al.(2022)Hui, Li, Hu, and Fu]{hui2022neural}
Ka-Hei Hui, Ruihui Li, Jingyu Hu, and Chi-Wing Fu.
\newblock Neural wavelet-domain diffusion for 3d shape generation.
\newblock In \emph{SIGGRAPH Asia 2022 Conference Papers}, pp.\  1--9, 2022.

\bibitem[Karras et~al.(2022)Karras, Aittala, Aila, and Laine]{karras2022elucidating}
Tero Karras, Miika Aittala, Timo Aila, and Samuli Laine.
\newblock Elucidating the design space of diffusion-based generative models.
\newblock \emph{Advances in neural information processing systems}, 35:\penalty0 26565--26577, 2022.

\bibitem[Kingma(2013)]{kingma2013auto}
Diederik~P Kingma.
\newblock Auto-encoding variational bayes.
\newblock \emph{arXiv preprint arXiv:1312.6114}, 2013.

\bibitem[Lei~Ba et~al.(2016)Lei~Ba, Kiros, and Hinton]{lei2016layer}
Jimmy Lei~Ba, Jamie~Ryan Kiros, and Geoffrey~E Hinton.
\newblock Layer normalization.
\newblock \emph{ArXiv e-prints}, pp.\  arXiv--1607, 2016.

\bibitem[Li et~al.(2023)Li, Tan, Zhang, Xu, Luan, Xu, Hong, Sunkavalli, Shakhnarovich, and Bi]{li2023instant3d}
Jiahao Li, Hao Tan, Kai Zhang, Zexiang Xu, Fujun Luan, Yinghao Xu, Yicong Hong, Kalyan Sunkavalli, Greg Shakhnarovich, and Sai Bi.
\newblock Instant3d: Fast text-to-3d with sparse-view generation and large reconstruction model.
\newblock \emph{arXiv preprint arXiv:2311.06214}, 2023.

\bibitem[Lin et~al.(2023)Lin, Gao, Tang, Takikawa, Zeng, Huang, Kreis, Fidler, Liu, and Lin]{lin2023magic3d}
Chen-Hsuan Lin, Jun Gao, Luming Tang, Towaki Takikawa, Xiaohui Zeng, Xun Huang, Karsten Kreis, Sanja Fidler, Ming-Yu Liu, and Tsung-Yi Lin.
\newblock Magic3d: High-resolution text-to-3d content creation.
\newblock In \emph{Proceedings of the IEEE/CVF Conference on Computer Vision and Pattern Recognition}, pp.\  300--309, 2023.

\bibitem[Liu et~al.(2024)Liu, Xu, Jin, Chen, Varma~T, Xu, and Su]{liu2024one}
Minghua Liu, Chao Xu, Haian Jin, Linghao Chen, Mukund Varma~T, Zexiang Xu, and Hao Su.
\newblock One-2-3-45: Any single image to 3d mesh in 45 seconds without per-shape optimization.
\newblock \emph{Advances in Neural Information Processing Systems}, 36, 2024.

\bibitem[Long et~al.(2024)Long, Guo, Lin, Liu, Dou, Liu, Ma, Zhang, Habermann, Theobalt, et~al.]{long2024wonder3d}
Xiaoxiao Long, Yuan-Chen Guo, Cheng Lin, Yuan Liu, Zhiyang Dou, Lingjie Liu, Yuexin Ma, Song-Hai Zhang, Marc Habermann, Christian Theobalt, et~al.
\newblock Wonder3d: Single image to 3d using cross-domain diffusion.
\newblock In \emph{Proceedings of the IEEE/CVF Conference on Computer Vision and Pattern Recognition}, pp.\  9970--9980, 2024.

\bibitem[Mittal et~al.(2022)Mittal, Cheng, Singh, and Tulsiani]{mittal2022autosdf}
Paritosh Mittal, Yen-Chi Cheng, Maneesh Singh, and Shubham Tulsiani.
\newblock Autosdf: Shape priors for 3d completion, reconstruction and generation.
\newblock In \emph{Proceedings of the IEEE/CVF Conference on Computer Vision and Pattern Recognition}, pp.\  306--315, 2022.

\bibitem[Morrison et~al.(2020)Morrison, Corke, and Leitner]{morrison2020egad}
Douglas Morrison, Peter Corke, and J{\"u}rgen Leitner.
\newblock Egad! an evolved grasping analysis dataset for diversity and reproducibility in robotic manipulation.
\newblock \emph{IEEE Robotics and Automation Letters}, 5\penalty0 (3):\penalty0 4368--4375, 2020.

\bibitem[Park et~al.(2019)Park, Florence, Straub, Newcombe, and Lovegrove]{park2019deepsdf}
Jeong~Joon Park, Peter Florence, Julian Straub, Richard Newcombe, and Steven Lovegrove.
\newblock Deepsdf: Learning continuous signed distance functions for shape representation.
\newblock In \emph{Proceedings of the IEEE/CVF conference on computer vision and pattern recognition}, pp.\  165--174, 2019.

\bibitem[Peebles \& Xie(2022)Peebles and Xie]{Peebles2022DiT}
William Peebles and Saining Xie.
\newblock Scalable diffusion models with transformers.
\newblock \emph{arXiv preprint arXiv:2212.09748}, 2022.

\bibitem[Petrov et~al.(2024)Petrov, Goyal, Thamizharasan, Kim, Gadelha, Averkiou, Chaudhuri, and Kalogerakis]{petrov2024gem3d}
Dmitry Petrov, Pradyumn Goyal, Vikas Thamizharasan, Vladimir Kim, Matheus Gadelha, Melinos Averkiou, Siddhartha Chaudhuri, and Evangelos Kalogerakis.
\newblock Gem3d: Generative medial abstractions for 3d shape synthesis.
\newblock In \emph{ACM SIGGRAPH 2024 Conference Papers}, pp.\  1--11, 2024.

\bibitem[Poole et~al.(2022)Poole, Jain, Barron, and Mildenhall]{poole2022dreamfusion}
Ben Poole, Ajay Jain, Jonathan~T Barron, and Ben Mildenhall.
\newblock Dreamfusion: Text-to-3d using 2d diffusion.
\newblock \emph{arXiv preprint arXiv:2209.14988}, 2022.

\bibitem[Qian et~al.(2023)Qian, Mai, Hamdi, Ren, Siarohin, Li, Lee, Skorokhodov, Wonka, Tulyakov, et~al.]{qian2023magic123}
Guocheng Qian, Jinjie Mai, Abdullah Hamdi, Jian Ren, Aliaksandr Siarohin, Bing Li, Hsin-Ying Lee, Ivan Skorokhodov, Peter Wonka, Sergey Tulyakov, et~al.
\newblock Magic123: One image to high-quality 3d object generation using both 2d and 3d diffusion priors.
\newblock \emph{arXiv preprint arXiv:2306.17843}, 2023.

\bibitem[Ren et~al.(2024)Ren, Huang, Zeng, Museth, Fidler, and Williams]{ren2024xcube}
Xuanchi Ren, Jiahui Huang, Xiaohui Zeng, Ken Museth, Sanja Fidler, and Francis Williams.
\newblock Xcube: Large-scale 3d generative modeling using sparse voxel hierarchies.
\newblock In \emph{Proceedings of the IEEE/CVF Conference on Computer Vision and Pattern Recognition}, pp.\  4209--4219, 2024.

\bibitem[Saharia et~al.(2022)Saharia, Chan, Saxena, Li, Whang, Denton, Ghasemipour, Gontijo~Lopes, Karagol~Ayan, Salimans, et~al.]{saharia2022photorealistic}
Chitwan Saharia, William Chan, Saurabh Saxena, Lala Li, Jay Whang, Emily~L Denton, Kamyar Ghasemipour, Raphael Gontijo~Lopes, Burcu Karagol~Ayan, Tim Salimans, et~al.
\newblock Photorealistic text-to-image diffusion models with deep language understanding.
\newblock \emph{Advances in neural information processing systems}, 35:\penalty0 36479--36494, 2022.

\bibitem[Shue et~al.(2023)Shue, Chan, Po, Ankner, Wu, and Wetzstein]{shue20233d}
J~Ryan Shue, Eric~Ryan Chan, Ryan Po, Zachary Ankner, Jiajun Wu, and Gordon Wetzstein.
\newblock 3d neural field generation using triplane diffusion.
\newblock In \emph{Proceedings of the IEEE/CVF Conference on Computer Vision and Pattern Recognition}, pp.\  20875--20886, 2023.

\bibitem[Siddiqui et~al.(2024)Siddiqui, Alliegro, Artemov, Tommasi, Sirigatti, Rosov, Dai, and Nie{\ss}ner]{siddiqui2024meshgpt}
Yawar Siddiqui, Antonio Alliegro, Alexey Artemov, Tatiana Tommasi, Daniele Sirigatti, Vladislav Rosov, Angela Dai, and Matthias Nie{\ss}ner.
\newblock Meshgpt: Generating triangle meshes with decoder-only transformers.
\newblock In \emph{Proceedings of the IEEE/CVF Conference on Computer Vision and Pattern Recognition}, pp.\  19615--19625, 2024.

\bibitem[Sun et~al.(2018)Sun, Wu, Zhang, Zhang, Zhang, Xue, Tenenbaum, and Freeman]{sun2018pix3d}
Xingyuan Sun, Jiajun Wu, Xiuming Zhang, Zhoutong Zhang, Chengkai Zhang, Tianfan Xue, Joshua~B Tenenbaum, and William~T Freeman.
\newblock Pix3d: Dataset and methods for single-image 3d shape modeling.
\newblock In \emph{Proceedings of the IEEE conference on computer vision and pattern recognition}, pp.\  2974--2983, 2018.

\bibitem[Tang et~al.(2023)Tang, Ren, Zhou, Liu, and Zeng]{tang2023dreamgaussian}
Jiaxiang Tang, Jiawei Ren, Hang Zhou, Ziwei Liu, and Gang Zeng.
\newblock Dreamgaussian: Generative gaussian splatting for efficient 3d content creation.
\newblock \emph{arXiv preprint arXiv:2309.16653}, 2023.

\bibitem[Vahdat \& Kautz(2020)Vahdat and Kautz]{vahdat2020nvae}
Arash Vahdat and Jan Kautz.
\newblock Nvae: A deep hierarchical variational autoencoder.
\newblock \emph{Advances in neural information processing systems}, 33:\penalty0 19667--19679, 2020.

\bibitem[Wang et~al.(2023)Wang, Du, Li, Yeh, and Shakhnarovich]{wang2023score}
Haochen Wang, Xiaodan Du, Jiahao Li, Raymond~A Yeh, and Greg Shakhnarovich.
\newblock Score jacobian chaining: Lifting pretrained 2d diffusion models for 3d generation.
\newblock In \emph{Proceedings of the IEEE/CVF Conference on Computer Vision and Pattern Recognition}, pp.\  12619--12629, 2023.

\bibitem[Wang \& Shi(2023)Wang and Shi]{wang2023imagedream}
Peng Wang and Yichun Shi.
\newblock Imagedream: Image-prompt multi-view diffusion for 3d generation.
\newblock \emph{arXiv preprint arXiv:2312.02201}, 2023.

\bibitem[Wang et~al.(2024)Wang, Lu, Wang, Bao, Li, Su, and Zhu]{wang2024prolificdreamer}
Zhengyi Wang, Cheng Lu, Yikai Wang, Fan Bao, Chongxuan Li, Hang Su, and Jun Zhu.
\newblock Prolificdreamer: High-fidelity and diverse text-to-3d generation with variational score distillation.
\newblock \emph{Advances in Neural Information Processing Systems}, 36, 2024.

\bibitem[Xiong et~al.(2024)Xiong, Wei, Zheng, Cao, Lian, and Wang]{xiong2024octfusion}
Bojun Xiong, Si-Tong Wei, Xin-Yang Zheng, Yan-Pei Cao, Zhouhui Lian, and Peng-Shuai Wang.
\newblock Octfusion: Octree-based diffusion models for 3d shape generation.
\newblock \emph{arXiv preprint arXiv:2408.14732}, 2024.

\bibitem[Xu et~al.(2023)Xu, Tan, Luan, Bi, Wang, Li, Shi, Sunkavalli, Wetzstein, Xu, et~al.]{xu2023dmv3d}
Yinghao Xu, Hao Tan, Fujun Luan, Sai Bi, Peng Wang, Jiahao Li, Zifan Shi, Kalyan Sunkavalli, Gordon Wetzstein, Zexiang Xu, et~al.
\newblock Dmv3d: Denoising multi-view diffusion using 3d large reconstruction model.
\newblock \emph{arXiv preprint arXiv:2311.09217}, 2023.

\bibitem[Yan et~al.(2022)Yan, Lin, Mitra, Lischinski, Cohen-Or, and Huang]{yan2022shapeformer}
Xingguang Yan, Liqiang Lin, Niloy~J Mitra, Dani Lischinski, Daniel Cohen-Or, and Hui Huang.
\newblock Shapeformer: Transformer-based shape completion via sparse representation.
\newblock In \emph{Proceedings of the IEEE/CVF Conference on Computer Vision and Pattern Recognition}, pp.\  6239--6249, 2022.

\bibitem[Yariv et~al.(2024)Yariv, Puny, Gafni, and Lipman]{yariv2024mosaic}
Lior Yariv, Omri Puny, Oran Gafni, and Yaron Lipman.
\newblock Mosaic-sdf for 3d generative models.
\newblock In \emph{Proceedings of the IEEE/CVF Conference on Computer Vision and Pattern Recognition}, pp.\  4630--4639, 2024.

\bibitem[Yi et~al.(2023)Yi, Fang, Wu, Xie, Zhang, Liu, Tian, and Wang]{yi2023gaussiandreamer}
Taoran Yi, Jiemin Fang, Guanjun Wu, Lingxi Xie, Xiaopeng Zhang, Wenyu Liu, Qi~Tian, and Xinggang Wang.
\newblock Gaussiandreamer: Fast generation from text to 3d gaussian splatting with point cloud priors.
\newblock \emph{arXiv preprint arXiv:2310.08529}, 2023.

\bibitem[Zeng et~al.(2022)Zeng, Vahdat, Williams, Gojcic, Litany, Fidler, and Kreis]{zeng2022lion}
Xiaohui Zeng, Arash Vahdat, Francis Williams, Zan Gojcic, Or~Litany, Sanja Fidler, and Karsten Kreis.
\newblock Lion: Latent point diffusion models for 3d shape generation.
\newblock \emph{arXiv preprint arXiv:2210.06978}, 2022.

\bibitem[Zhang \& Wonka(2024)Zhang and Wonka]{zhang2024functional}
Biao Zhang and Peter Wonka.
\newblock Functional diffusion.
\newblock In \emph{Proceedings of the IEEE/CVF Conference on Computer Vision and Pattern Recognition}, pp.\  4723--4732, 2024.

\bibitem[Zhang et~al.(2022)Zhang, Nie{\ss}ner, and Wonka]{zhang20223dilg}
Biao Zhang, Matthias Nie{\ss}ner, and Peter Wonka.
\newblock 3dilg: Irregular latent grids for 3d generative modeling.
\newblock \emph{Advances in Neural Information Processing Systems}, 35:\penalty0 21871--21885, 2022.

\bibitem[Zhang et~al.(2023)Zhang, Tang, Nie\ss{}ner, and Wonka]{vecset}
Biao Zhang, Jiapeng Tang, Matthias Nie\ss{}ner, and Peter Wonka.
\newblock 3dshape2vecset: A 3d shape representation for neural fields and generative diffusion models.
\newblock \emph{ACM Trans. Graph.}, 42\penalty0 (4), July 2023.
\newblock ISSN 0730-0301.
\newblock \doi{10.1145/3592442}.
\newblock URL \url{https://doi.org/10.1145/3592442}.

\bibitem[Zhang et~al.(2024{\natexlab{a}})Zhang, Wang, Zhang, Qiu, Pang, Jiang, Yang, Xu, and Yu]{clay}
Longwen Zhang, Ziyu Wang, Qixuan Zhang, Qiwei Qiu, Anqi Pang, Haoran Jiang, Wei Yang, Lan Xu, and Jingyi Yu.
\newblock Clay: A controllable large-scale generative model for creating high-quality 3d assets.
\newblock \emph{ACM Trans. Graph.}, 43\penalty0 (4), July 2024{\natexlab{a}}.
\newblock ISSN 0730-0301.
\newblock \doi{10.1145/3658146}.
\newblock URL \url{https://doi.org/10.1145/3658146}.

\bibitem[Zhang et~al.(2024{\natexlab{b}})Zhang, Wang, Zhang, Qiu, Pang, Jiang, Yang, Xu, and Yu]{zhang2024clay}
Longwen Zhang, Ziyu Wang, Qixuan Zhang, Qiwei Qiu, Anqi Pang, Haoran Jiang, Wei Yang, Lan Xu, and Jingyi Yu.
\newblock Clay: A controllable large-scale generative model for creating high-quality 3d assets.
\newblock \emph{ACM Transactions on Graphics (TOG)}, 43\penalty0 (4):\penalty0 1--20, 2024{\natexlab{b}}.

\bibitem[Zhao et~al.(2024)Zhao, Liu, Chen, Zeng, Wang, Cheng, Fu, Chen, Yu, and Gao]{zhao2024michelangelo}
Zibo Zhao, Wen Liu, Xin Chen, Xianfang Zeng, Rui Wang, Pei Cheng, Bin Fu, Tao Chen, Gang Yu, and Shenghua Gao.
\newblock Michelangelo: Conditional 3d shape generation based on shape-image-text aligned latent representation.
\newblock \emph{Advances in Neural Information Processing Systems}, 36, 2024.

\bibitem[Zheng et~al.(2023)Zheng, Pan, Wang, Tong, Liu, and Shum]{zheng2023locally}
Xin-Yang Zheng, Hao Pan, Peng-Shuai Wang, Xin Tong, Yang Liu, and Heung-Yeung Shum.
\newblock Locally attentional sdf diffusion for controllable 3d shape generation.
\newblock \emph{ACM Transactions on Graphics (ToG)}, 42\penalty0 (4):\penalty0 1--13, 2023.

\bibitem[Zheng et~al.(2024)Zheng, Pan, Guo, Tong, and Liu]{zheng2024mvd}
Xin-Yang Zheng, Hao Pan, Yu-Xiao Guo, Xin Tong, and Yang Liu.
\newblock Mvd\^{} 2: Efficient multiview 3d reconstruction for multiview diffusion.
\newblock In \emph{ACM SIGGRAPH 2024 Conference Papers}, pp.\  1--11, 2024.

\bibitem[Zhou \& Jacobson(2016)Zhou and Jacobson]{zhou2016thingi10k}
Qingnan Zhou and Alec Jacobson.
\newblock Thingi10k: A dataset of 10,000 3d-printing models.
\newblock \emph{arXiv preprint arXiv:1605.04797}, 2016.

\end{thebibliography}
\bibliographystyle{iclr2025_conference}

\appendix
\section{Data preprocessing}
The data preprocessing is based on~\citep{zhang20223dilg}.
\subsection{Volume points sampling.}
We sample volume points uniformly in the bounding sphere.
\begin{lstlisting}[language=Python]
N_vol = 250000
vol_points = np.random.randn(N_vol, 3)
vol_points = vol_points / np.linalg.norm(vol_points, axis=1)[:, None] * np.sqrt(3)
vol_points = vol_points * np.power(np.random.rand(N_vol), 1./3)[:, None]
\end{lstlisting}

\subsection{Near points sampling}
The near-surface points are obtained by sampling Gaussian-jittered surface points.
\begin{lstlisting}[language=Python]
N_near = 125000
# surface_points: N_near x 3
near_points = [
    surface_points + np.random.normal(scale=0.005, size=(N_near, 3)),
    surface_points + np.random.normal(scale=0.05, size=(N_near, 3)),
]
near_points = np.concatenate(near_points)
\end{lstlisting}
\section{Data augmentations}
\paragraph{Random axis scaling.} The augmentation is from~\citep{zhang20223dilg}. We randomly sample a scaling factor for each axis which ranges from [0.75, 1.25].
\paragraph{Unit sphere normalization.} We normalize each mesh to a unit sphere, i.e., the max point norm of the point clouds is 1.
\begin{lstlisting}[language=Python]
# v: vertices n x 3
v = v - (v.max(axis=0) + v.min(axis=0)) / 2
distances = np.linalg.norm(v, axis=1)
scale = 1 / np.max(distances)
v *= scale
\end{lstlisting}
\paragraph{Random rotations.} We apply random rotations during the training of the autoencoder, 
\begin{equation}
    \rmR(\alpha, \beta, \gamma) = \begin{bmatrix}
        \cos\alpha & -\sin\alpha & 0\\
        \sin\alpha & \cos\alpha & 0 \\
        0 & 0 & 1\\
    \end{bmatrix}
    \begin{bmatrix}
        \cos\beta & 0 & \sin\beta\\
        0 & 1 & 0 \\
        -\sin\beta & 0 & \cos\beta \\
    \end{bmatrix}
    \begin{bmatrix}
        1 & 0 & 0 \\
        0 & \cos\gamma & -\sin\gamma \\
        0 & \sin\gamma & \cos\gamma \\
    \end{bmatrix},
\end{equation}
where $\alpha$, $\beta$, and $\gamma$ are yaw, pitch, and roll, respectively. Our meshes are firstly normalized into a unit sphere. Thus after the random rotations, the models will still be inside of a unit sphere.
\section{Regularization}
The proposed regularization (see~\cref{tab:ftl_ltf}) is implemented with layer normalization (PyTorch code).
\begin{lstlisting}[language=Python]
# network definition
self.ftl_proj = nn.Linear(x_dim, z_dim)
self.ftl_norm = nn.LayerNorm(dims, elementwise_affine=False, eps=1e-6)
# network forward
z = self.ftl_norm(self.ftl_proj(x))
\end{lstlisting}

\section{Training time query points sampling}
In the previous work~\citep{zhang20223dilg}, the sampling strategy is uniformly sampling 1024 points in the bounding volume during training. We found this is not working on Objaverse. Since lots of meshes have very thin structures, this strategy will cause no inside points to be sampled during training. This heavily imbalenced data classficiation severely affects the occupancy loss.

We propose the following solution. In each iteration, we make sure half of the points have positive labels and the other half have negative labels.

\section{Training loss}
The loss is binary cross entropy as in previous work~\citep{zhang20223dilg}. Formally, we have
\begin{equation}
\gL = \mathbb{E}_{\mathbf{p}\in\mathbb{R}^3}\left[\mathrm{BCE}\left(
        \mathcal{\hat{O}}(\mathbf{p}), \mathcal{O}(\mathbf{p})
    \right)\right].
\end{equation}
In practice, we use the empircal loss
\begin{equation}
\mathbb{E}_{\mathbf{p}\in\gQ^{\text{vol}}}\left[\mathrm{BCE}\left(
        \mathcal{\hat{O}}(\mathbf{p}), \mathcal{O}(\mathbf{p})
    \right)\right] + 0.1 \cdot \mathbb{E}_{\mathbf{p}\in\gQ^{\text{near}}}\left[\mathrm{BCE}\left(
        \mathcal{\hat{O}}(\mathbf{p}), \mathcal{O}(\mathbf{p})
    \right)\right].
\end{equation}
Here, $\gQ^{\text{vol}}$ is the set of volume query points, and $\gQ^{\text{near}}$ is the set of near-surface query points.
\section{Diffusion}
We use the formulation EDM~\citep{karras2022elucidating} for the diffusion models. The inference/sampling algorithm is also taken from the paper.

\section{Latents analysis}
We analyze how latents are affecting the final reconstruction. The latents are partially replaced by standard Gaussian noise (this is because our latents are also zero mean and unit variance). We show the visual results in~\cref{fig:noise}.

\definecolor{gencol}{RGB}{0,0,255}
\definecolor{noisecol}{RGB}{255,69,0}
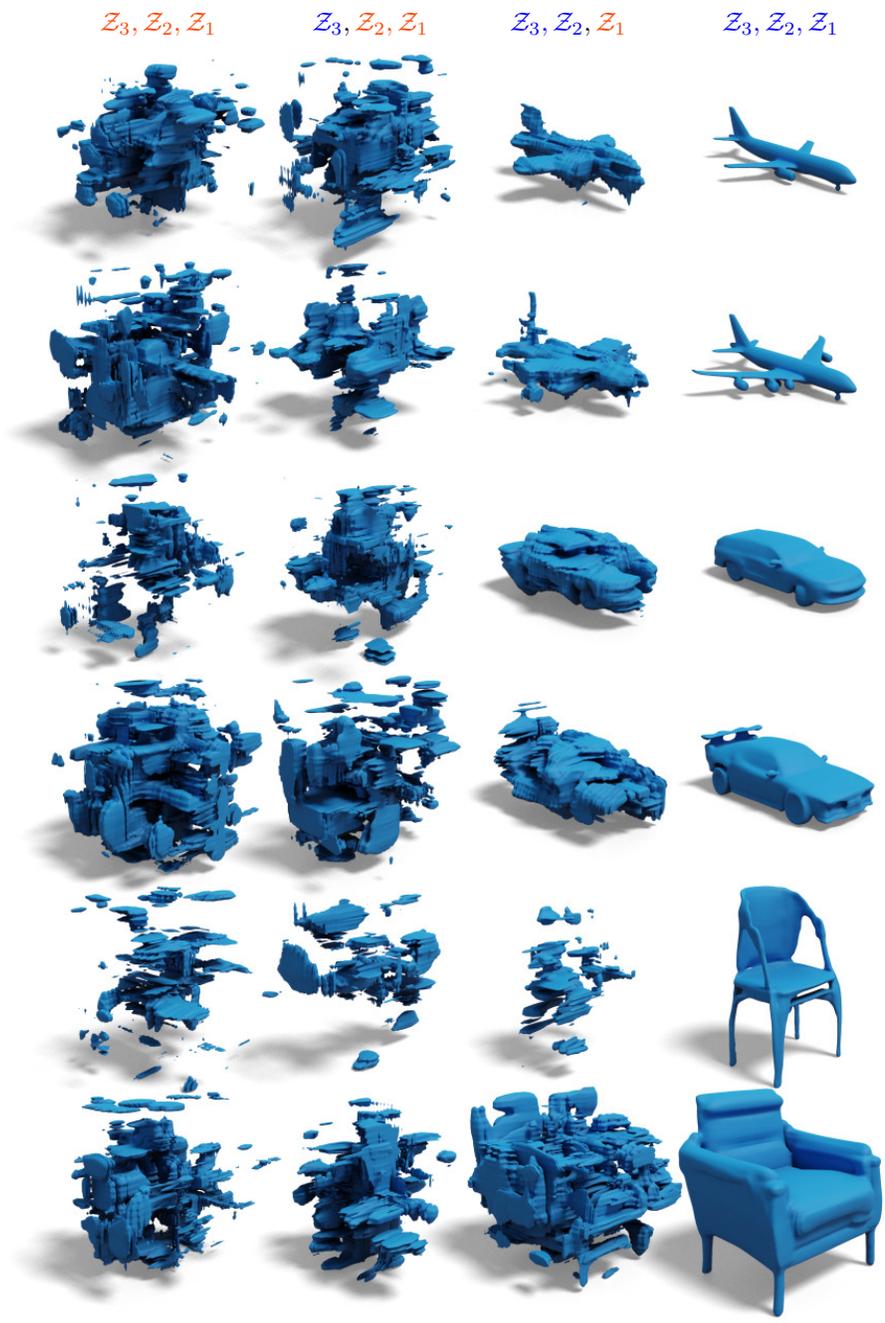
\begin{figure}
    \centering
    \begin{overpic}[trim={0cm 0cm 0cm 0cm},clip,
    width=0.95\linewidth,
    grid=false]{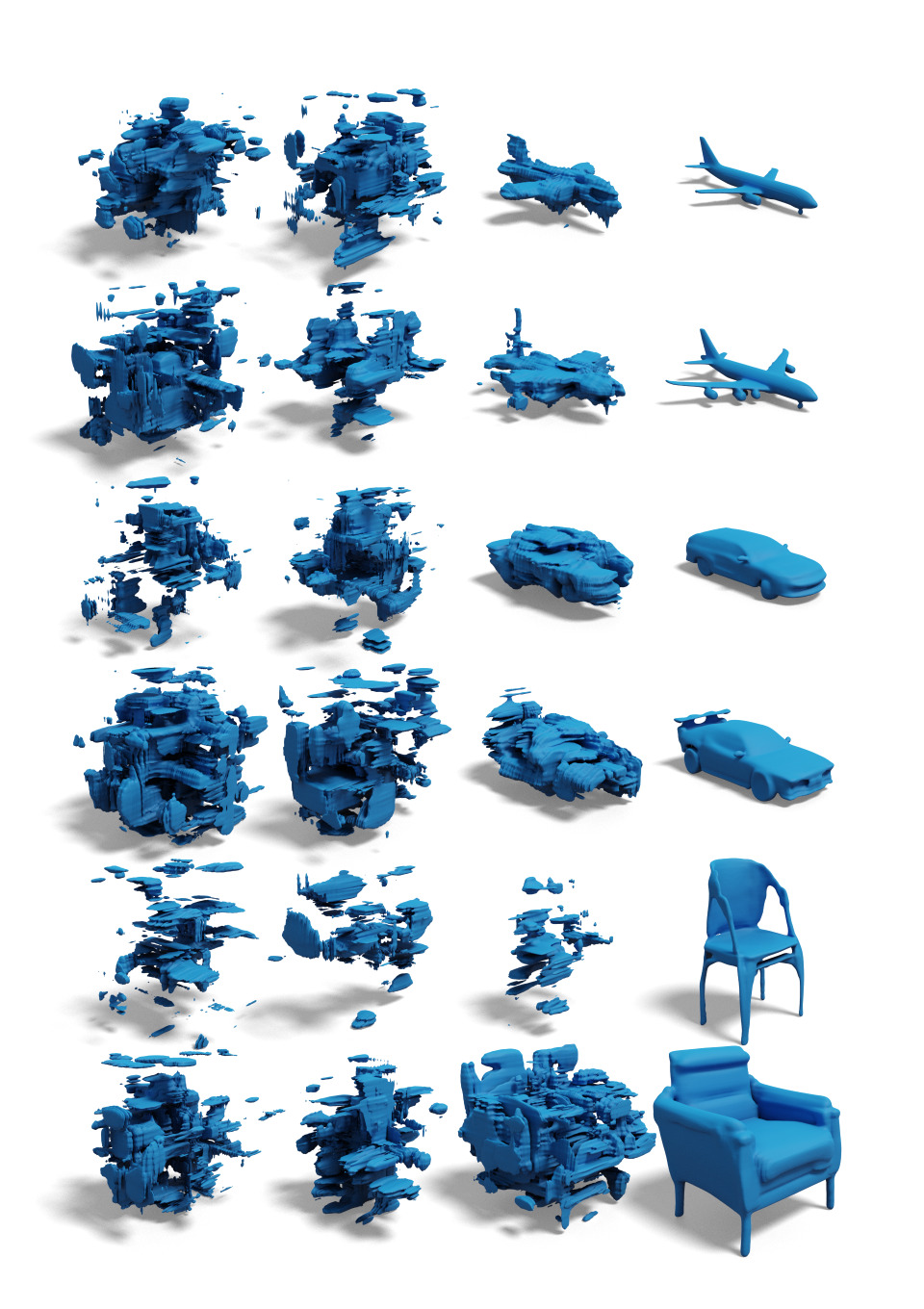}
        \put(9,95){$\textcolor{noisecol}{\gZ_3, \gZ_2, \gZ_1}$}
        \put(24,95){$\textcolor{gencol}{\gZ_3}, \textcolor{noisecol}{\gZ_2, \gZ_1}$}
        \put(38,95){$\textcolor{gencol}{\gZ_3, \gZ_2}, \textcolor{noisecol}{\gZ_1}$}
        \put(53,95){$\textcolor{gencol}{\gZ_3, \gZ_2, \gZ_1}$}
    \end{overpic}
    \vspace{-10pt}
    \caption{Latent with red color \textcolor{noisecol}{$\gZ$} means it is replaced by Gaussian noise. Latent with blue color \textcolor{gencol}{$\gZ$} means it is generated with the diffusion models.}
    \label{fig:noise}
\end{figure}

\end{document}